\documentclass{egpubl}
\ifxetex
  
\fi
\usepackage{pg2026}

\SpecialIssuePaper

\CGFccby

\usepackage[T1]{fontenc}
\usepackage{dfadobe}
\usepackage{float}
\usepackage{amsfonts}
\usepackage{amssymb}
\usepackage{amsmath}
\usepackage{cite}
\BibtexOrBiblatex

\electronicVersion
\PrintedOrElectronic

\usepackage{graphicx}

\usepackage{egweblnk}
\usepackage{cleveref}

\usepackage{amsfonts}
\newcommand{\tablestyle}[2]{\setlength{\tabcolsep}{#1}\renewcommand{\arraystretch}{#2}}
\usepackage{booktabs}
\usepackage{multirow}
\usepackage[table]{xcolor}
\usepackage{capt-of} % for \captionof outside floating environments
\usepackage{microtype}

\newcommand{\changed}[1]{#1}

\newif\ifincludeappendix
\includeappendixtrue
\usepackage{amsmath}
\usepackage{pifont}

\title[Dream4D]%
      {Dream4D: Lifting Camera-Controlled I2V towards Spatiotemporally Consistent 4D Generation}

\author[X. Liu et al.]
{\parbox{\textwidth}{\centering
Xiaoyan Liu$^{1,\dagger}$,
Kangrui Li$^{2,\dagger}$,
Jiaxin Liu$^{3,\dagger}$,
Yuehao Song$^{4}$,
Yujie Xin$^{4,*}$
}
\\
{\parbox{\textwidth}{\centering
$^{1}$The Chinese University of Hong Kong\\
$^{2}$The Hong Kong Polytechnic University\\
$^{3}$The University of New South Wales\\
$^{4}$Huazhong University of Science and Technology
}
}
}
\begin{document}

% uncomment for using teaser
% \teaser{
%  \includegraphics[width=0.9\linewidth]{eg_new}
%  \centering
%   \caption{New EG Logo}
% \label{fig:teaser}
%}

\maketitle
\begingroup
\renewcommand{\thefootnote}{\dagger}
\footnotetext{These authors contributed equally to this work.}
\renewcommand{\thefootnote}{*}
\footnotetext{Corresponding author.}
\endgroup
\hypersetup{%
  pdftitle={Dream4D: Lifting Camera-Controlled I2V towards Spatiotemporally Consistent 4D Generation},%
  pdfauthor={Xiaoyan Liu, Kangrui Li, Jiaxin Liu, Yuehao Song, Yujie Xin},%
  pdfkeywords={4D generation; camera-controlled image-to-video synthesis; dynamic scene reconstruction; video diffusion models},%
  linkcolor=black,%
  citecolor=black,%
  urlcolor=black%
}
%-------------------------------------------------------------------------
\begin{abstract}
Generating spatiotemporally consistent 4D content requires joint modeling of high-fidelity spatial details and realistic temporal dynamics.
However, existing methods typically lack world knowledge learned from expansive 4D observations.
As a result, generated scenes often struggle to maintain pose and temporal consistency in complex dynamic environments, leading to degraded generation quality.
To address these challenges, we propose \textbf{Dream4D}, a novel 4D generation framework that integrates spatiotemporal priors from video generation with explicit sequential camera control.
Our method adopts a three-stage pipeline.
We first employ a vision--language model (VLM) to determine the camera trajectory that best captures the intended view from the input image and the instruction text.
Next, a pose-conditioned diffusion model generates geometrically consistent image sequences along this trajectory.
Finally, we use an advanced 4D generator to reconstruct a coherent and persistent 4D representation conditioned on the generated video.
\textbf{Dream4D} jointly leverages the rich temporal priors of video diffusion models and the geometric awareness of reconstruction models to achieve high-quality image-to-4D generation. Extensive experiments demonstrate that our approach significantly outperforms existing methods in both pose and temporal consistency.
% \vspace{-1em}
%-------------------------------------------------------------------------
%  ACM CCS 1998
%  (see https://www.acm.org/publications/computing-classification-system/1998)
% \begin{classification} % according to https://www.acm.org/publications/computing-classification-system/1998
% \CCScat{Computer Graphics}{I.3.3}{Picture/Image Generation}{Line and curve generation}
% \end{classification}
%-------------------------------------------------------------------------
%  ACM CCS 2012

%The tool at \url{http://dl.acm.org/ccs.cfm} can be used to generate
% CCS codes.
%Example:


\end{abstract}  
%-------------------------------------------------------------------------

\section{Introduction}

\label{sec:intro}

\begin{figure*}[t]
    \centering
    \includegraphics[width=0.92\textwidth]{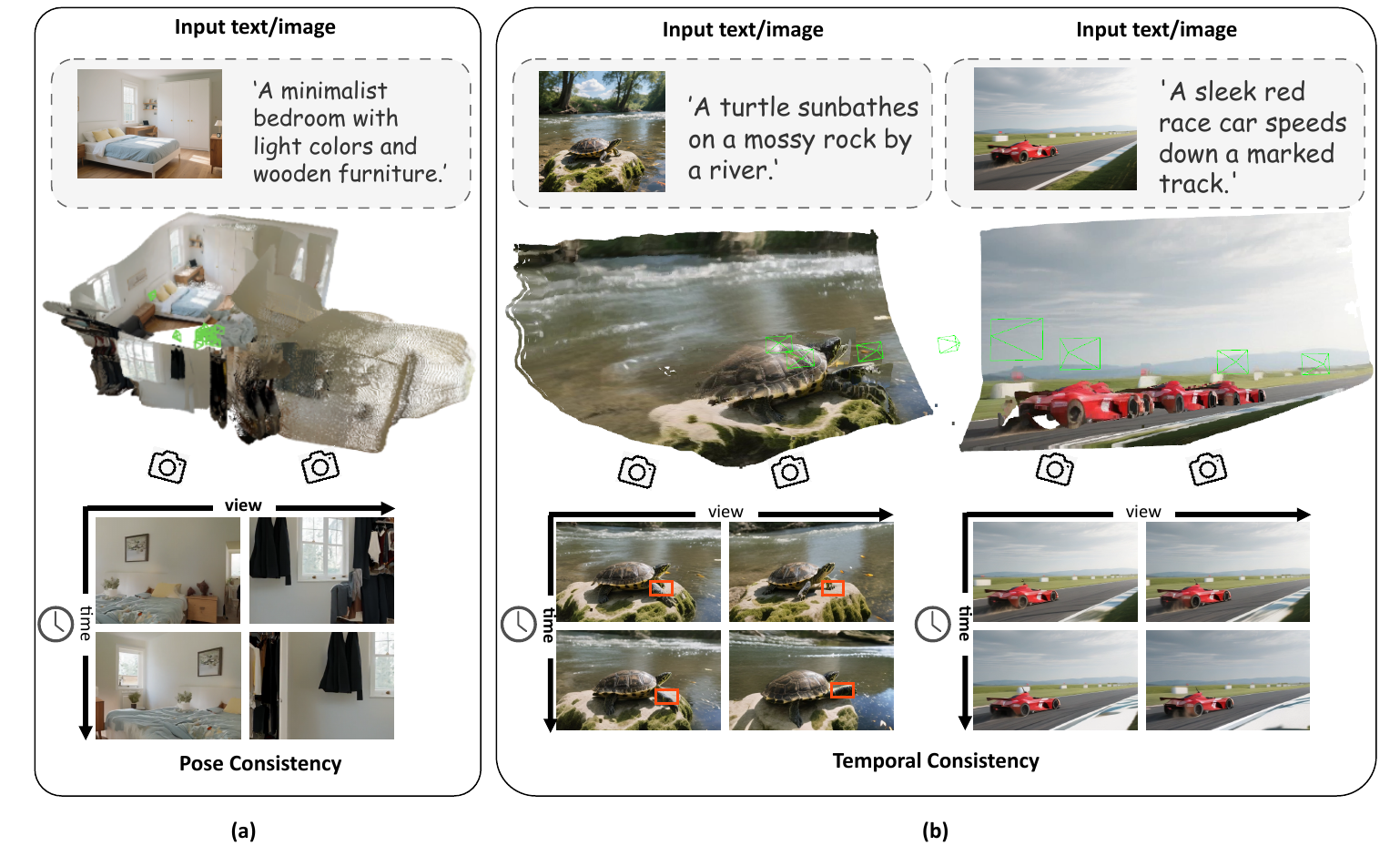}
    \vspace{-0.55em}
    \caption{\textbf{Spatiotemporally consistent 4D generation.} Our method achieves excellent consistency in both pose and temporal dimensions.}
    \label{fig:main_result}
\end{figure*}

4D content generation extends traditional 3D reconstruction by introducing a temporal dimension~\cite{5}, enabling the creation of dynamic scenes that evolve over time and giving rise to a broad spectrum of application demands across numerous domains~\cite{1,2,3,4}. The central challenge of this task lies in the joint modeling of high-fidelity spatial representations and physically plausible temporal dynamics~\cite{6}, a challenge that becomes particularly acute in the presence of large camera motions or highly dynamic scenes. We decompose this challenge into two dimensions: \textit{temporal consistency}, which requires that the appearance and geometry of objects in a dynamic scene exhibit smooth, flicker-free continuous evolution along the temporal axis~\cite{xiangli2022bungeenerf}; and \textit{pose consistency}, which demands that the 3D geometry remain accurate and coherent across different viewpoints, ensuring distortion-free rendering from arbitrary camera poses. However, existing 4D generation methods~\cite{D-nerf, Flownet3d, Diffcloth, yang2023diffusion} predominantly focus on the modeling of spatiotemporal variations themselves, offering insufficient capability for jointly guaranteeing both forms of consistency, especially in complex scenes.

Recent advances in video generation models have demonstrated their remarkable capacity to learn spatiotemporal priors from large-scale video data, capturing coherent motion patterns and consistent pose relationships~\cite{voleti2024sv3d, ren2025gen3c, bar2024lumiere}. Integrating such priors into 4D generation represents a promising avenue for enhancing consistency, yet a fundamental difficulty persists: the effective transfer of these priors necessitates the imposition of geometric constraints aligned with the scene context (such as specific camera trajectories) to maintain coherence across viewpoints and temporal frames. Different scenes impose markedly different trajectory requirements, a panoramic room scene calls for a sweeping rotational scan, whereas a racing scenario demands lateral translational tracking, implying that trajectory planning cannot be predetermined as fixed but must be jointly guided by visual observations and textual input to achieve adaptive alignment with the specific scene context.

To address this challenge, we propose Dream4D, a holistic framework that decomposes 4D generation into three core stages. Given an image and its accompanying textual description, Dream4D first leverages a Vision-Language Model (VLM) to interpret scene semantics and plan a camera pose trajectory aligned with the scene context. A pose-conditioned video diffusion (VD) model then generates a dynamically coherent image sequence that encapsulates rich spatiotemporal priors, providing high-quality visual guidance for subsequent reconstruction. Finally, the spatiotemporal information within the generated video is employed as constraints to elevate the 2D sequence into a 4D representation through 4D reconstruction. Our core insight resides in the synergistic integration of cross-stage priors: the VLM furnishes high-level semantic intent, the VD module translates this intent into smooth and coherent spatiotemporal signals, and the 4D reconstruction module acts as an enforcer of global geometric constraints, rigorously correcting the consistency of the visual priors produced by the VD output. Each of the three stages fulfills a distinct role while mutually reinforcing the others, rendering robust and high-fidelity 4D generation possible at a level that surpasses what can be achieved by 2D diffusion models or standalone 3D reconstruction pipelines alone.

As illustrated in Figure~\ref{fig:main_result}, Dream4D achieves a favorable balance between pose consistency (Figure~\ref{fig:main_result}(a)) and temporal consistency (Figure~\ref{fig:main_result}(b)). Quantitative experiments further corroborate this qualitative observation: Dream4D achieves a significant improvement in pose consistency (a relative translation error (RTE) reduction of 18.2\% over the online baseline), along with a notable improvement in temporal consistency ($\Delta$mPSNR improvement of 2.935 over the baseline).

Our contributions are summarized as follows:

\vspace{-0.75mm}
\begin{itemize}
\item \textbf{Unified pose-conditioned 4D generation.} We propose Dream4D, the first image-to-4D generation framework with explicit camera trajectory guidance, achieving coherent 4D content creation by employing a pose-conditioned video diffusion model as the carrier of spatiotemporal priors.
\item \textbf{VLM-based scene-aware pose trajectory planning.} We leverage a Vision-Language Model to interpret scene semantics and generate context-aligned viewpoint trajectories, translating semantic intent into physically coherent visual sequences via the video diffusion model and enabling semantically controllable 4D generation.
\item \textbf{Superior spatiotemporal consistency and generation quality.} Comprehensive evaluation demonstrates that Dream4D simultaneously achieves high pose consistency (18.2\% RTE reduction) and temporal consistency (2.935 $\Delta$mPSNR improvement) in 4D generation, validating the efficacy of the cross-stage prior synergistic strategy.
\end{itemize}

%-------------------------------------------------------------------------

\section{Related Work}
\label{sec:rw}

\begin{figure*}[t]
    \centering
    \includegraphics[width=0.94\linewidth]{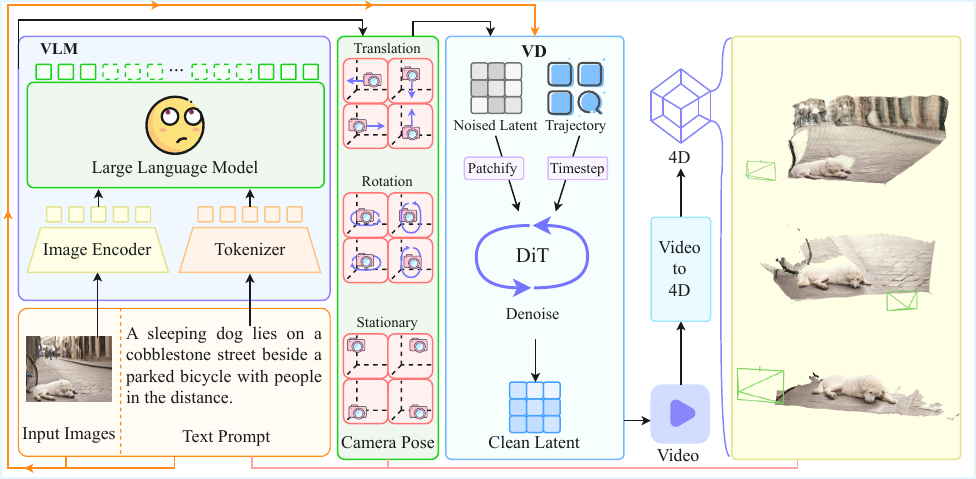}
    \caption{\textbf{Pipeline Overview.} Our method consists of three stages, i.e., vision language model (VLM), video diffusion model (VD), and video-to-4D generation. Given the input images and the text prompt containing the scene information and the requirements for camera control, the VLM transmits them to a conditional representation that describes the camera pose sequence to guide the subsequent generation process. The camera poses are categorized into translation, rotation, and stationary for fine-grained control. The VLM predicted camera pose trajectory is further encoded into a trajectory condition and fed into a Diffusion Transformer (DiT)~\cite{peebles2023scalablediffusionmodelstransformers} based video diffusion model (VD). The generated videos are then passed to a 4D generator, which transforms them into the final 4D output.}
    \label{fig:pipline}
\end{figure*}

\subsection{Camera-Controlled Video Generation}
Recent research on camera-controlled video generation primarily focuses on two interconnected technical pathways, i.e., 3D-aware generation and contextual understanding. 
SeVA~\cite{SeVA} incorporates camera motion via epipolar constraints and spatial attention mechanisms to maintain multi-view coherence in static scenes. In contrast, CamI2V~\cite{cami2v} achieves dynamic content synthesis by conditioning on both camera trajectories and text prompts, establishing SE(3)-equivariant representations to improve multi-view consistency.
%In contrast, CamI2V~\cite{cami2v} achieves dynamic content synthesis by conditioning on both camera trajectories and text prompts, establishing SE(3)-equivariant representations for improved multi-view consistency.%
Recent efforts have further decoupled camera control from content generation ~\cite{wang2024motionctrl} or injected explicit 3D priors into the diffusion process \cite{voleti2024sv3d, ren2025gen3c, yu2024viewcrafter}. Despite these architectural refinements, the basic formulation remains largely unchanged. \changed{In the camera-controlled I2V methods considered here, trajectories primarily serve as synthesis-time conditions and are not explicitly reused as geometric constraints in a downstream persistent 4D reconstruction stage.} As a result, these methods implicitly treat video synthesis as the terminal objective rather than as an intermediate observation process for spatial-temporal reconstruction. Synthesized frames are never fed into a geometric reconstruction pipeline, leaving no persistent 4D representation and no mechanism to enforce metric geometric consistency through explicit pose constraints.

% Furthermore, recent work has contributed to controllable and spatiotemporally consistent video generation through advanced adapter modules~\cite{he2024cameractrl}, epipolar attention with reference frame conditioning~\cite{xu2024camco}, and framework optimization~\cite{zhao2024motiondirector,avetisyan2024scenescript}.

% These methods demonstrate advanced temporal consistency, while overlook explicit 3D geometric accuracy and exhibit physically implausible interactions and limited generalization.

\subsection{4D Reconstruction}

4D reconstruction aims to recover dynamic 3D scenes that evolve over time.
Early methods rely on Structure-from-Motion (SfM)~\cite{schonberger2016structure, fisher2021colmap} and SLAM~\cite{orbslam} pipelines.
They are effective for sparse static scenes but often struggle with dense dynamic content.
%The field is revolutionized by neural rendering paradigms.
Neural rendering paradigms have revolutionized the field.
Implicit representations, such as Deformable NeRF~\cite{D-nerf} and its variants\cite{nerf, park2021hypernerf}, enable high-fidelity modeling of dynamic scenes, while explicit methods, particularly 4D extensions of 3D Gaussian Splatting (3DGS)~\cite{wu2024recent, chen2024survey, luiten2024dynamic, wang2024shape}, achieve real-time performance. Currently, the ``3R'' framework family (Cut3R~\cite{cut3r}, DUSt3R~\cite{wang2024dust3r}, MASt3R~\cite{leroy2024grounding}) unifies the recovery of structures and the reasoning of the scene. Extensions such as MonST3R ~\cite{zhang2024monst3r} and SPann3R ~\cite{wang20253d} have adapted this family to dynamic scenes, introducing motion-aware geometric reasoning. 
These methods demonstrate advanced fidelity, temporal consistency, and rendering speed of dynamic scene reconstruction. However, these methods often face a trade-off between visual quality and geometric completeness, particularly when dealing with sparse or noisy inputs.
More fundamentally, they are designed to reconstruct from captured video, not to synthesize and then reconstruct from a single image under an explicitly planned camera trajectory. The input observation density is treated as given, rather than actively generated.

\subsection{3D/4D Generation}

3D and 4D content generation expands significantly beyond static object creation to complex dynamic scenes, yet controllability remains the central unresolved problem. On the static front, InstantMesh ~\cite{xu2024instantmesh} and Zero123++ ~\cite{shi2023zero123++} produce multi-view geometry from a single image but offer no temporal dynamics or camera control.  
WonderJourney~\cite{yu2024wonderjourney} and WonderWorld~\cite{yu2025wonderworld} generate explorable scenes from single images or text prompts, demonstrating compelling scene transitions and rapid generation capabilities.
However, these approaches struggle to maintain geometric completeness and temporal consistency in expansive, dynamic scenes.
DreamGaussian4D ~\cite{ren2023dreamgaussian4d} and Animate124 ~\cite{zhao2023animate124} generate dynamic scenes from single images by combining 3D priors with temporal diffusion, but neither offers explicit camera trajectory control.
In the realm of dynamic generation, CAT4D~\cite{wu2025cat4d} integrates multi-view diffusion with deformable 3DGS to enhance dynamic object rendering.
Subsequent works such as MoDGS~\cite{liu2024modgs} and MoSca~\cite{lei2024mosca} improve the practicality by enabling the reconstruction of casually captured monocular videos.
However, it is challenging for these methods to achieve long-term spatiotemporal consistency in complex environments.
In contrast, our approach tackles complex scene-level 4D generation by coupling a VLM-guided video diffusion process with a 4D neural representation module, achieving consistent results over long temporal horizons.

\section{Method}
\label{sec:method}

Dream4D takes as input a single image $I$ and a text instruction $c$, and produces a 4D representation that remains stable under prescribed camera motion while exhibiting plausible temporal evolution. The problem is inherently constrained: the generated scene must respect the semantic intent of the prompt, follow a physically meaningful camera path, and maintain a persistent geometry across time. We address these requirements with a three-stage framework. A vision--language model first infers a scene-appropriate camera trajectory from the image--text pair (Sec.~\ref{sec:trajectory_selection}). A pose-conditioned video diffusion model then generates a temporally coherent image sequence along this trajectory (Sec.~\ref{sec:video_generation}). Finally, a pose-aware 4D reconstruction module lifts the generated sequence into a consistent spatiotemporal representation (Sec.~\ref{sec:4dgen}). An overview of our method is shown in Fig.~\ref{fig:pipline}.

% An overview of our method is shown in Fig.~\ref{fig:pipline}.
% Our method contains three stages: VLM-based pose trajectory planning (Sec.~\ref{sec:trajectory_selection}), pose-conditioned video generation (Sec.~\ref{sec:video_generation}) and 4D generation via pose-aware reconstruction (Sec.~\ref{sec:4dgen}). 
% Together, they form a synergistic system designed for cascading error compensation (\cref{subsec:ssec}).

\subsection{VLM-Based Pose Trajectory Planning}
\label{sec:trajectory_selection}

Dynamic 4D scene generation requires camera motion to be semantically consistent with the scene's logic.
To this end, we plan the camera pose based on the semantics extracted from the input image and the corresponding description.

Given the input image $I$ and textual instruction $c$, the VLM predicts a sequence of discrete camera-motion tokens $\mathbf{a}_{1:T}=(a_1,a_2,\ldots,a_T)$ from a predefined action space:

\begin{equation}
    \mathcal{A} = \{\mathrm{stationary}, \mathrm{translation}_{d,\lambda}, \mathrm{rotation}_{d,\lambda}\}
\end{equation}

where $d \in \{up, down, left, right\}$ denotes the spatial direction and $\lambda \in \{subtle, moderate, wide\}$ controls the motion amplitude.\changed{Concretely, the three motion-amplitude levels correspond to rotation angles of $\{2^\circ, 6^\circ, 12^\circ\}$ and
translation magnitudes of $\{0.05, 0.15, 0.30\}$ in the normalized scene coordinate system, respectively. Each primitive is smoothly interpolated over its assigned frame interval.}
The model is trained to maximize the conditional log-likelihood:

\begin{equation}
\log p_\phi(\mathbf{a}_{1:T} \mid I,c) = \sum_{t=1}^{T} \log p_\phi(a_t \mid I,c,\mathbf{a}_{1:t-1}),
\end{equation}

where $\phi$ denotes the VLM parameters. During inference, we decode the most probable trajectory as

\begin{equation}
\hat{\mathbf{a}}_{1:T} = \arg\max_{\mathbf{a}_{1:T}} p_\phi(\mathbf{a}_{1:T} \mid I,c).
\end{equation}

Specifically, the VLM outputs a sequence of discrete motion tokens representing the intended camera primitives. These tokens are then mapped to a parameterized SE(3) camera path $\{\changed{\mathbf{C}_t}\}_{t=1}^{T}$ through a deterministic parameterization process, where each world-to-camera pose $\changed{\mathbf{C}_t}=[\mathbf{R}_t\mid\mathbf{t}_t]\in\mathrm{SE}(3)$ is defined by a rotation matrix $\mathbf{R}_t\in\mathrm{SO}(3)$ and a translation vector $\mathbf{t}_t\in\mathbb{R}^{3}$ at time $t$. Thus, $a_t$ denotes a discrete VLM motion token, whereas $\changed{\mathbf{C}_t}$ denotes the continuous camera pose at frame $t$. This token-to-SE(3) mapping is deterministic and non-learned: the VLM selects the motion primitives, and their conversion to continuous poses follows a fixed analytical rule without trainable parameters.

\begin{table*}[t]
    \centering
    %\footnotesize  % 建议移除或注释掉，否则即使填满，字也会很小
    \tablestyle{2.5pt}{1.2}
    \caption{\textbf{Evaluation on Pose Consistency} on Sintel~\cite{butler2012naturalistic}, TUM-dynamic~\cite{sturm2012benchmark}, ScanNet~\cite{dai2017scannet}    and Bonn~\cite{palazzolo2019refusion} datasets. The best and second-best results are denoted in \textbf{bold} and \underline{underlined}, respectively. The 4D generator in our method is built upon an online approach and our method achieves more competitive results than optimization methods and the best overall performance among all online methods.\changed{``--'' indicates that the result is neither reported in the original paper nor obtainable from official checkpoints under this evaluation setting.}}
    \label{tab:table8}
    % 关键修改：使用 tabular*，并添加 @{\extracolsep{\fill}}
    \begin{tabular*}{\linewidth}{@{\extracolsep{\fill}} l c c c c c c c c c c c c}
    \toprule
    \multirow{2}{*}{\centering Method} 
    & \multicolumn{3}{c}{Sintel} 
    & \multicolumn{3}{c}{TUM-dynamics} 
    & \multicolumn{3}{c}{ScanNet} 
    & \multicolumn{3}{c}{Bonn}  \\
    \cmidrule(lr){2-4} \cmidrule(lr){5-7} \cmidrule(lr){8-10} \cmidrule(lr){11-13}
    & ATE\,$\downarrow$ & RTE\,$\downarrow$ & RRE\,$\downarrow$ 
    & ATE\,$\downarrow$ & RTE\,$\downarrow$ & RRE\,$\downarrow$  
    & ATE\,$\downarrow$ & RTE\,$\downarrow$ & RRE\,$\downarrow$  
    & ATE\,$\downarrow$ & RTE\,$\downarrow$ & RRE\,$\downarrow$  \\
    \midrule
    \multicolumn{13}{c}{Optimization-Based Method} \\
    \midrule
    CasualSAM~\cite{zhang2022structure} & \underline{0.141} & \textbf{0.035} & \textbf{0.615} & \underline{0.045} & 0.020 & \underline{0.841} & 0.158 & 0.034 &     1.618 & - & - & -  \\
    DUSt3R~\cite{wang2024dust3r} & 0.417 & 0.250 & 5.796 & 0.127 & 0.062 & 3.099 & 0.081 & 0.028 & 0.784 & - & - & -  \\
    MASt3R~\cite{leroy2024grounding} & 0.185 & 0.060 & \underline{1.496} & \textbf{0.038} &\textbf{0.012} & \textbf{0.448} & \underline{0.078} &     \underline{0.020} & \textbf{0.475} & - & - & -  \\
    MonST3R~\cite{zhang2024monst3r} & \textbf{0.111} & \underline{0.044} & 0.869 & 0.098 & \underline{0.019} & 0.935 & \textbf{0.077} & \textbf{0.018} &    \underline{0.529} & 0.129 & 0.018 & 0.863  \\
    \midrule
    \multicolumn{13}{c}{Online Methods} \\
    \midrule
    SPann3R~\cite{wang20253d} & 0.329 & 0.110 & 4.471 & 0.056 & 0.021 & 0.591 & \underline{0.096} & 0.023 & 0.661 & 0.074 & 0.022 &  0.638 \\
    CUT3R~\cite{cut3r} & \underline{0.213} & \underline{0.066} & \underline{0.621} & \underline{0.046} & \underline{0.015} & \underline{0.473} & 0.099 &    \underline{0.022} & \underline{0.600} & \underline{0.042} & \underline{0.013} & \textbf{0.629}  \\
    \rowcolor{gray!20}
    Ours & \textbf{0.210} & \textbf{0.054} & \textbf{0.619} & \textbf{0.044} & \textbf{0.014} & \textbf{0.445} & \textbf{0.082} & \textbf{0.019} & \textbf{0.504} & \textbf{0.038} & \textbf{0.008} & \underline{0.635}  \\
    \bottomrule
    \end{tabular*}
\end{table*}

\vspace{-3mm}
{\setlength{\textfloatsep}{6pt}% space between text and floats at top/bottom
 \setlength{\intextsep}{6pt}% space for in-text floats
\begin{table}[t]
    \centering
   % \small  % 或 
    % \scriptsize
    \normalsize
    \renewcommand{\arraystretch}{1.1}
    \caption{\textbf{Temporal Consistency Evaluation.} Our method achieves superior performance over baselines.}
    \label{tab:comparison}
    \begin{tabular}{@{}lccc@{}}
    \toprule
    \textbf{Method} & \textbf{mPSNR$\uparrow$} & \textbf{mSSIM$\uparrow$} & \textbf{mLPIPS$\downarrow$} \\
    \midrule
    MegaSAM~\cite{li2024_megasam} & 17.625 & 0.601 & 0.207 \\
    Shape-of-Motion~\cite{wang2024shape} & 16.68 & 0.630 & 0.332 \\
    Cut3R~\cite{cut3r} & 14.69 & 0.543 & 0.341 \\
    % DynIBaR~\cite{li2023dynibar} & 12.67 & 0.495 & 0.579 \\
    \rowcolor{gray!20}
    \textbf{Ours} & \textbf{20.56} & \textbf{0.702} & \textbf{0.170} \\
    \bottomrule
    \end{tabular}
\end{table}}

\subsection{Pose-Conditioned Video Generation}
\label{sec:video_generation}

% To incorporate world knowledge as spatiotemporal priors, we employ a pose-conditioned VD model. The model is conditioned on camera pose parameters to ensure coherent temporal evolution and structural stability in the generated sequences. 
% The effectiveness of this method is closely related to the quality and semantic richness of the camera pose itself. 
% Therefore, we employ the VLM as an intelligent interpreter to translate high-level instructions or contextual understanding into precise and semantically meaningful camera poses.
% The camera pose embedding produced by the VLM encodes a geometric prior that directly guides the spatiotemporal exploration of the scene. 

To incorporate world knowledge as spatiotemporal priors, we employ a VD model conditioned on camera pose parameters, which ensures coherent temporal evolution and structural stability.  Crucially, the effectiveness of this approach hinges on the quality and semantic richness of the camera poses themselves. Therefore, we utilize a VLM as an intelligent interpreter to translate high-level instructions or contextual understanding into precise and semantically meaningful camera poses. The resulting pose embedding from the VLM encodes a geometric prior that directly guides the spatiotemporal exploration of the scene.
% It should be noted that if output trajectory embeddings are not specifically enhanced for downstream tasks and are transferred directly, it can lead to an inefficient dissipation of model capacity.  

% This shortcoming often manifests at later stages, forcing the model to rely on computationally expensive reconstruction and resampling operations for compensation.

%It results in a generated video sequence that is essentially a geometrically consistent dynamic view sequence sampled along a predetermined trajectory.

% To transform this discrete motion command into a continuous, visually coherent video sequence, we leverage a video diffusion model conditioned on both semantic content and explicit camera poses. 
%It results in a generated video sequence that is essentially a geometrically consistent dynamic view sequence sampled along a predetermined trajectory.

% (e.g., \textit{ turn left})
The pose embedding produced is then assigned to a parameterized SE(3) camera path $\{\changed{\mathbf{C}_t}\}_{t=1}^{T}$. This path is not learned, but analytically defined based on the motion type (e.g., circular orbit around the scene center or linear pan along the horizontal axis).

In practice, the diffusion model (e.g., a DiT-based architecture~\cite{peebles2023scalablediffusionmodelstransformers}) processes two distinct conditioning signals. The text instruction $Instr$ and reference image $I$ supply semantic information but carry no metric geometric meaning. We inject these implicit conditions through standard cross-attention:

\begin{equation}\label{eq:semantic_attention}
\operatorname{Attn}_{\mathrm{sem}}(Q,K_{\mathrm{sem}},V_{\mathrm{sem}})=\operatorname{Softmax}\left(\frac{QK_{\mathrm{sem}}^{\top}}{\sqrt{d_k}}\right)V_{\mathrm{sem}},
\end{equation}

where $K_{\mathrm{sem}}$ and $V_{\mathrm{sem}}$ derive from the joint text-image embedding. The camera pose $\changed{\mathbf{C}_t}\in\mathrm{SE}(3)$, by contrast, constitutes an explicit geometric constraint that fixes the viewpoint in world coordinates. To prevent semantic ambiguity from interfering with spatial precision, we process the pose through a separate geometric pathway. We first encode $\changed{\mathbf{C}_t}$ into a high-dimensional embedding $e_t^{\mathrm{pose}}=\mathrm{MLP}_{\mathrm{pose}}\bigl(\operatorname{vec}(\changed{\mathbf{C}_t})\bigr)$, then modulate the DiT blocks via adaptive layer normalization:

\begin{equation}\label{eq:adaln}
\mathrm{AdaLN}(z_{\tau,t},e_t^{\mathrm{pose}})=\gamma(e_t^{\mathrm{pose}})\odot z_{\tau,t}+\beta(e_t^{\mathrm{pose}}),
\end{equation}

where $t$ indexes the video frame, $\tau$ indexes the diffusion denoising step, and $z_{\tau, t}$ is the intermediate latent associated with frame $t$ at denoising step $\tau$. The functions $\gamma$ and $\beta$ predict affine parameters from the pose embedding.

Formally, the generation process factorizes autoregressively across time:

% uses $\mathbf{C}_t$ as an additional input in each denoising step, typically through cross-attention or adaptive layer normalization, ensuring that the generated frames respect the intended camera motion. Based on these conditions, the video generation process conditioned on the initial image $I$, the instruction $Instr$, and the pose sequence $\{\mathbf{C}_t\}$, is formally defined by:
\begin{equation}
    p(V)=\prod_{t=1}^{T}p_{\theta}(V_t\mid I, c, \mathbf{C}_t),
\end{equation}

where $V = \{V_t\}_{t=1}^T$ is the generated video. Training minimizes the standard noise-prediction objective:

% The generation process is guided by a denoising objective where the model learns to predict the noise added to a clean video latent. This process is conditioned on the initial image $I$, the instruction text $Instr$, and the planned camera pose sequence $\mathbf{C}_t$. The Diffusion Model Denoising Objectives are:

\begin{equation}
\mathcal{L}_{\mathrm{DM}} = \mathbb{E}_{z_0, \epsilon, \tau} \left[ \left\| \epsilon - \epsilon_\theta(z_\tau, \tau, I, c, \{C_t\}_{t=1}^T) \right\|_2^2 \right],
\end{equation}

\vspace{-3mm}
where $\epsilon_{\theta}$ is the noise predicted by the model parameterized with $\theta$, $\epsilon$ is the sampled noise, $z_\tau$ is the noisy video latent at diffusion step $\tau$, and ${C_t}^T_{t=1}$ is the frame-wise camera trajectory.

To further enhance geometric consistency and temporal smoothness during the generation process, we introduce additional constraints in the denoising process of the diffusion model. Adjacent views must also agree geometrically. We therefore add a warping consistency term to the denoising objective. For consecutive frames $V_t$ and $V_{t+1}$, we define the source-to-target relative pose as $\Delta C_t = C_{t+1} C_t^{-1}$ and penalize the photometric discrepancy between $V_{t+1}$ and the forward-warped $V_t$:

% In addition to the original reconstruction loss, we have added a geometric consistency loss based on camera pose $\mathcal{L}_{\text {geo }}$. For a small pose change $\Delta \mathbf{C}$, the corresponding change should satisfy the following:
\begin{equation}
    \mathcal{L}_{\mathrm{geo}}=\sum_t\left\|V_{t+1}-\operatorname{Warp}\left(V_t,\changed{\Delta\mathbf{C}_t}\right)\right\|_1.
    \label{equ:3}
\end{equation}

% Here, $\Delta \mathbf{C}$ represents the camera pose transformation matrix connecting two frames, and $V_{t}$ and $V_{t+1}$ respectively represent the video frames generated by the model at time steps $t$ and $t+1$. 

The warping function is defined by camera intrinsics and the relative pose transformation.

\begin{equation}
\operatorname{Warp}(V_t, \Delta C_t)(p') = V_t\!\left(\pi\!\left(\mathbf{K} \Delta C_t^{-1} \Pi^{-1}\!\left(p', D_{t+1}(p')\right)\right)\right),
\end{equation}

where $p'$ is a target pixel in frame $V_{t+1}$, $\Pi^{-1}$ back-projects it using the target-frame depth $D_{t+1}(p')$ and camera intrinsics, and $\pi$ denotes perspective projection. Since $\Delta C_t$ maps source-camera coordinates to target-camera coordinates, $\Delta C_t^{-1}$ maps the target-frame point back to the source camera for sampling $V_t$

Notably, our camera control is not restricted to rigid categories. Instead, we employ a dual-conditioning strategy that treats VLM tokens as parameterized primitives while leveraging textual prompts for implicit semantic guidance.

% At the same time, we introduce three key components:

% \noindent    
% \textbf{Pose Correction Layer}: Minimizes reprojection errors between predicted and rendered views via differentiable rendering, refining pose alignment during generation.

% \noindent  
% \textbf{Occlusion-Sensitive Attention Masking}: Dynamically adjusts attention weights to suppress features from disoccluded regions during viewpoint transitions.

% \noindent  
% \textbf{Depth-Guided Temporal Super-Resolution}: Enhances frame resolution while preserving depth coherence across time, preventing temporal flickering.

% The output is a temporally coherent video sequence that simulates a virtual camera moving along the VLM-selected path. Crucially, this sequence constitutes a dynamic view sequence: it provides a \textit{structured, pose-aligned observation sequence} that serves as direct input for 4D reconstruction, effectively exposing the diffusion model's internal spatio-temporal priors under controlled viewing conditions.

\subsection{4D Generation via Pose-Aware Reconstruction}
\label{sec:4dgen}

% This section presents our framework for reconstructing temporally coherent 4D scenes from the generated multi-view video sequences and their associated camera trajectories. 

The generated video provides dense visual evidence along a planned camera trajectory. We reconstruct a temporally coherent 4D representation from the generated frames and their associated camera poses. This stage converts pose-conditioned video priors into an explicit spatiotemporal scene representation that supports consistent rendering across both time and viewpoint.

\noindent
\textbf{Initialization of the Temporal Structure}. Given the generated video $V=\{V_t\}_{t=1}^{T}$ and the corresponding camera poses $\{\changed{\mathbf{C}_t}\}_{t=1}^{T}$, we estimate per-frame monocular depth maps $D_t$ and inter-frame optical flow fields $F_{t\rightarrow t+1}$ to establish geometric and motion cues across views.
% \textbf{Temporal Structure Initialization}. Given the synthesized video $V \in R^{T\times H\times W \times 3}$ and camera poses $\left\{\mathbf{C}_{t}\right\}^T_{t=1}$, we first decompose each frame into per-frame depth maps $D_t$ and optical flow fields $F_{t \rightarrow {t+1}}$, aligning them using known camera poses.
The monocular depth network $\operatorname{DepthNet}_\theta$ predicts metric depth maps $D_t$:

\begin{equation}
    D_t=\operatorname{DepthNet}(V_t), \quad D_t\in R^{H\times W}.
\end{equation}

% These are aligned to a common coordinate frame using camera poses:

% \begin{equation}
%     \mathbf{D}_{t}^{\text {aligned }}=\mathbf{C}_{t}^{-1} \circ \mathbf{D}_{t} \circ \mathbf{C}_{t}
% \end{equation}

% Where $\circ$ represents the composite operation of transformation. 

To fuse multi-view geometric information, we back-project each depth map $D_t$ into a 3D point cloud $\mathcal{X}_t$ using the inverse camera projection:

\begin{equation}
\mathcal{X}_t=\{\changed{\mathbf{C}_t^{-1}}\cdot\Pi^{-1}(u,v,D_t(u,v))\mid(u,v)\in\Omega\},
\label{equ:5}
\end{equation}
where $\Pi^{-1}$ denotes the intrinsic back-projection, and $\Omega$ is the image domain. Thus, all point clouds $\{\mathcal{X}_t\}$ are expressed in a shared world coordinate system defined by the initial pose $\changed{\mathbf{C}_1}$.

\noindent
\textbf{Pose-Conditioned 4D Reconstruction}.
We build on CUT3R~\cite{cut3r}, which regresses per-frame pointmaps in a shared world frame and maintains temporal persistence through recurrent scene-state tokens. Our modification injects explicit camera
pose conditioning into this prediction process (see Appendix~\cref{app:A} for details). Specifically, we define a spatiotemporal feature field:

\begin{equation}
    \mathbf{Q}(\mathbf{x},t)=\operatorname{Transformer}_{\psi}\left(\gamma(\mathbf{x}),h(t),\changed{\mathbf{C}_t}\right),
    \label{equ:6}
\end{equation}
where $\gamma(\mathbf{x})$ denotes the positional encoding of the 3D coordinates $\mathbf{x}$, $h(t)$ is a temporal embedding, and $\changed{\mathbf{C}_t}$ is the camera extrinsic at time $t$.

Here, $\changed{\mathbf{C}_t}$ is first processed through an MLP to extract high-dimensional pose embeddings, which are then fused with positional and temporal embeddings via Transformer cross-attention layers.
% This architectural adaptation allows the model to correlate local geometry with the viewing angle, resolving depth ambiguities and ensuring structural persistence during long-range camera translations.

The Transformer cross-attention layers explicitly correlate camera poses with local geometry.

\begin{figure}[!t]
  \centering
  \includegraphics[width=\linewidth]{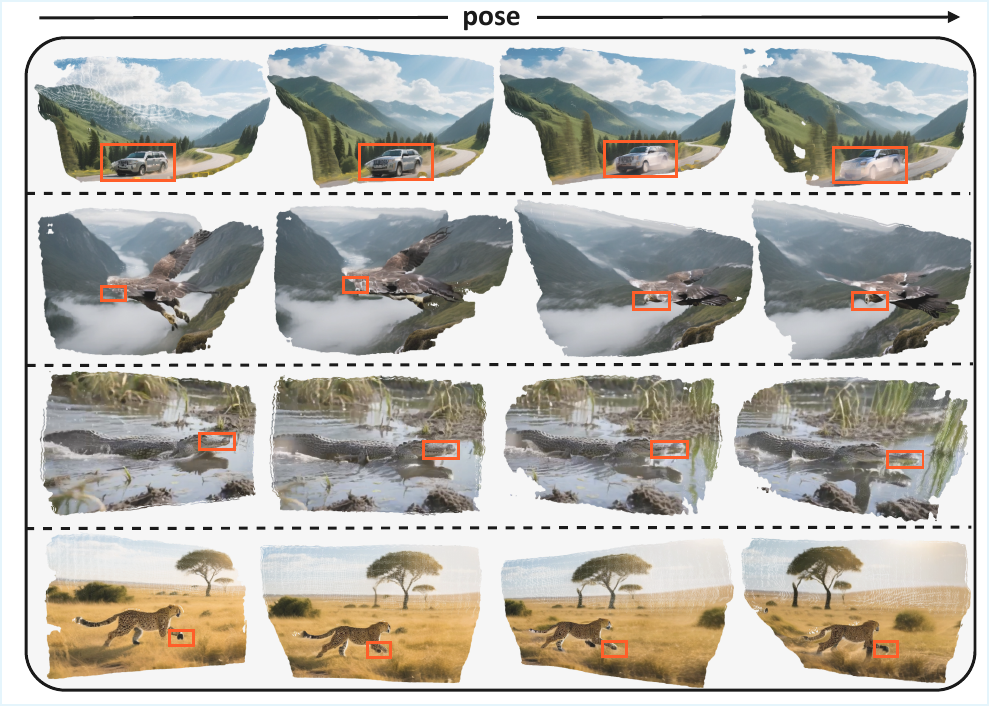}
  \caption{\textbf{Qualitative Evaluation of Pose Consistency.} This visualization compares the structural integrity and positional coherence of a target object across sequential frames under varying camera poses. Red boxes denote target object states.}
  \label{fig:dynamic-demo}
\end{figure}

\begin{table}[!t]
  \centering
  \caption{\textbf{Ablation Study on VLM.} Comparing the video generation quality of full model (denoted as w/ VLM) against two ablated variants: VD with random trajectory input (denoted as Random) and VD without trajectory input (denoted as Empty), using five complementary metrics.}
  \label{tab:ablation}
  \vspace{2pt}
  \tablestyle{0.5pt}{1.2}
  \begin{tabular}{lcccccc}
  \toprule
  \textbf{Variant} & \textbf{mPSNR (dB) $\uparrow$} & \textbf{mSSIM $\uparrow$}     & \textbf{mLPIPS $\downarrow$} & \textbf{CLIP-T $\uparrow$} & \textbf{CLIP-I $\uparrow$} \\
  \midrule
  w/ VLM & \textbf{21.07} & \textbf{0.7387} & \textbf{0.1222} & \textbf{0.3548} & \textbf{0.9472} \\
  \rowcolor{gray!10}
  Random & 17.37 & 0.6997 & 0.1659 & 0.3394 & 0.9276 \\
  \rowcolor{gray!20}
  $\Delta$ & -3.7 & -0.039 & +0.0437 & -0.0154 & -0.0196 \\
  \rowcolor{gray!10}
  Empty & 16.98 & 0.5415 & 0.3216 & 0.3337 & 0.9241 \\
  \rowcolor{gray!20}
  $\Delta$ &  -4.09 &  -0.1972 &  +0.1994 & -0.0211 & -0.0231 \\
  \bottomrule
  \end{tabular}
\end{table}

\subsection{Joint Optimization Strategy}

% Although Dream4D is organized into three functional stages, the full framework is optimized under a unified training objective that couples semantic trajectory planning, pose-conditioned video synthesis, and pose-aware 4D reconstruction. This joint formulation ensures that the predicted camera trajectory is not only semantically plausible, but also beneficial for downstream video generation and temporally coherent 4D reconstruction.

Although Dream4D follows a three-stage design, its components are optimized with a coordinated training strategy to maintain consistency across trajectory planning, pose-conditioned video synthesis, and 4D reconstruction. Each module is first initialized with supervision appropriate to its own prediction space: semantic motion primitives for the VLM planner, pose-conditioned denoising for the video diffusion model, and rendering-based reconstruction for the 4D generator. The camera trajectory provides the common geometric variable that links these stages. After stage-wise convergence, we fine-tune the lightweight pose encoders, conditioning layers, and reconstruction adapters under shared consistency objectives, while keeping the large pretrained backbones fixed or partially frozen. This strategy avoids unstable gradients through discrete trajectory selection and diffusion sampling, yet still aligns the full pipeline through a shared pose representation and downstream reconstruction constraints.

\begin{figure}[!t]
  \centering
  \includegraphics[width=\linewidth]{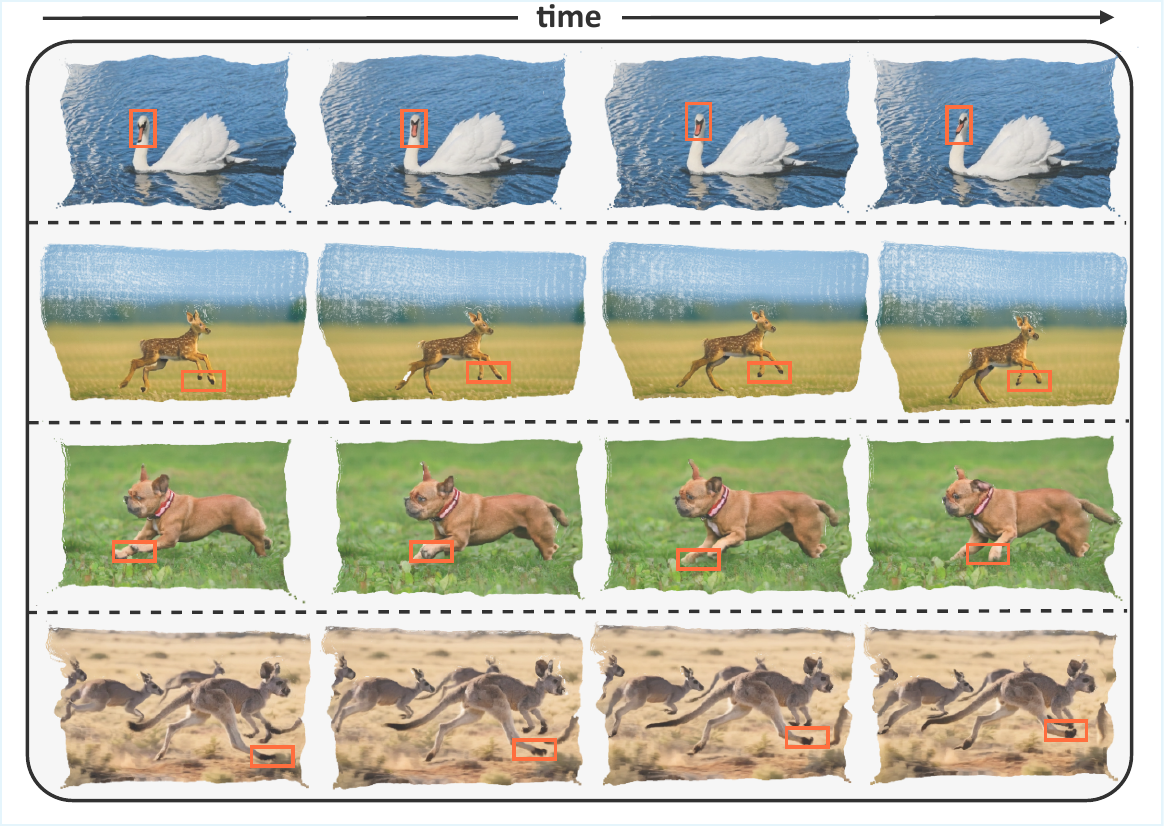}
  \caption{\textbf{Qualitative Evaluation of Temporal Consistency.} This visualization compares the smoothness and coherence of object motion across sequential frames under fixed camera poses. Red boxes denote target object states.}
  \label{fig:static-demo}
\end{figure}

\begin{table}[!t]
    \centering
    \caption{\textbf{Extra Ablation Study on VLM on the VBench benchmark.} The advanced performance of Dream4D on the VBench benchmark further demonstrates the impact of the VLM on video generation quality. Imaging Quality (IQ) evaluates visual defects frame-by-frame using a discriminative model. Temporal Flickering (TF) quantifies inter-frame stability in a deep feature space. Aesthetic Quality (AQ) assesses overall visual appeal through an aesthetic model. 
    % Together, these metrics comprehensively validate the contribution of the VLM. 
}
    \label{tab:ablation4}
    \tablestyle{12pt}{1.1}
    \begin{tabular}{lp{1.3cm}p{1.3cm}p{1.3cm}}
    \toprule
    \textbf{Variant} & \textbf{IQ$\uparrow$} & \textbf{TF$\uparrow$} & \textbf{AQ$\uparrow$} \\
    \midrule
    w/ VLM  & \textbf{71.22} & \textbf{0.9897} & \textbf{0.7307} \\
    \rowcolor{gray!10}
    Random  & 67.47 & 0.9776 & 0.6591 \\
    \rowcolor{gray!20}
    $\Delta$ & -3.75 & -0.0121 & -0.0716 \\
    \rowcolor{gray!10}
    Empty  & 61.57 & 0.9523 & 0.6251 \\
    \rowcolor{gray!20}
    $\Delta$ &  -9.65 &  -0.0374 &  -0.1056 \\
    \bottomrule
    \end{tabular}
\end{table}

Given an input image $I$, text instruction $c$, target video sequence $V^{*}=\{V_t^{*}\}_{t=1}^{T}$, and the corresponding camera trajectory $\changed{\mathbf{C}^{*}=\{\mathbf{C}_t^{*}\}_{t=1}^{T}}$, the VLM predicts a discrete motion sequence $\mathbf{a}_{1:T}$, which is deterministically converted into the continuous pose trajectory $\changed{\hat{\mathbf{C}}_{\phi}=\{\mathbf{C}_t\}_{t=1}^{T}}$. The video diffusion module then generates a pose-conditioned video sequence $V=\{V_t\}_{t=1}^{T}$, which is passed to the 4D reconstruction module to recover a dynamic scene representation. The overall objective is defined as:
% \begin{equation}\label{eq:total_loss}
% \begin{split}
% \mathcal{L}_{\mathrm{total}} &=
% \lambda_{\mathrm{plan}}\mathcal{L}_{\mathrm{plan}}
% + \lambda_{\mathrm{diff}}\mathcal{L}_{\mathrm{diff}}
% + \lambda_{\mathrm{geo}}\mathcal{L}_{\mathrm{geo}} \\
% &\quad + \lambda_{\mathrm{rec}}\mathcal{L}_{\mathrm{rec}}
% + \lambda_{\mathrm{temp}}\mathcal{L}_{\mathrm{temp}},
% \end{split}
% \end{equation}
\begin{equation}\label{eq:total_loss_compact}
\begin{split}
\min_{\phi,\theta,\psi}\,
\mathbb{E}_{(I,c,V^{*},\changed{\mathbf{C}^{*}})\sim\mathcal{D}}
\Big[
\lambda_{\mathrm{plan}}\mathcal{L}_{\mathrm{plan}}(\phi)
& + \lambda_{\mathrm{DM}}\mathcal{L}_{\mathrm{DM}}(\theta;\changed{\hat{\mathbf{C}}_{\phi}}) \\
+ \lambda_{\mathrm{geo}}\mathcal{L}_{\mathrm{geo}}(\theta;\changed{\hat{\mathbf{C}}_{\phi}})
+ \lambda_{\mathrm{rec}}\mathcal{L}_{\mathrm{rec}}(\psi;\hat{V}_{\theta},\changed{\hat{\mathbf{C}}_{\phi}})
&+ \lambda_{\mathrm{temp}}\mathcal{L}_{\mathrm{temp}}(\psi)
\Big],
\end{split}
\end{equation}
where $\phi$, $\theta$, and $\psi$ denote the trainable parameters of the VLM trajectory planner, video diffusion model, and 4D reconstruction module, respectively; $\changed{\hat{\mathbf{C}}_{\phi}}$ is the camera trajectory predicted and parameterized by the VLM; and $\hat{V}_{\theta}$ is the video generated by the pose-conditioned video diffusion (VD) model. Each term constrains a distinct level of the generation process. The planning loss $\mathcal{L}_{\mathrm{plan}}$ supervises the VLM trajectory prediction using the autoregressive next-token objective introduced in Sec.~\ref{sec:trajectory_selection}. The diffusion loss $\mathcal{L}_{\mathrm{DM}}$ and the geometric consistency term $\mathcal{L}_{\mathrm{geo}}$ are defined in Sec.~\ref{sec:video_generation}.

\begin{table}[!t]
    \centering
    \tablestyle{3pt}{1.1}
    \caption{\textbf{Quantitative Ablation Study on VD.} The world knowledge prior from the VD improves 4D generation quality.}
    \label{tab:ablation2}
    \begin{tabular}{lccp{1.7cm}}
    \toprule
    \textbf{Variant} & \textbf{mPSNR (dB)$\uparrow$} & \textbf{mSSIM$\uparrow$} & \textbf{mLPIPS$\downarrow$} \\
    \midrule
    w/ Dynamic Prior  & \textbf{19.78} & \textbf{0.6686} & \textbf{0.1220} \\
    w/o Dynamic Prior  & 18.37 & 0.6361 & 0.1841 \\
    \rowcolor{gray!20}
    $\Delta$ & -1.41 & -0.0325 & +0.0621 \\
    \midrule
    w/ Static Prior  & 13.35 & 0.6550 & 0.2996 \\
    w/o Static Prior  & 12.56 & 0.5779 & 0.3461 \\
    \rowcolor{gray!20}
    $\Delta$ &  -0.79 &  -0.0771 &  +0.0465 \\
    \bottomrule
    \end{tabular}
\end{table}

% follows the standard denoising objective and optimizes the pose-conditioned video generator to recover clean video latents from noised inputs:
% \begin{equation}\label{eq:diff_loss}
% \begin{split}
% \mathcal{L}_{\mathrm{DM
% }} =
% \mathbb{E}_{z_0,\epsilon,\tau}\Bigg[
% \Big\| &\epsilon -
% \epsilon_{\theta}(z_{\tau},\tau,I,c,\{\mathbf{C}_t\}_{t=1}^{T})
% \Big\|_2^2 \Bigg],
% \end{split}
% \end{equation}
% where $z_0$ denotes the clean video latent, $z_{\tau}$ is the noised latent at diffusion step $\tau$, and $\epsilon_{\theta}$ is the predicted noise.
% The geometric consistency term $\mathcal{L}_{\mathrm{geo}}$ enforces local agreement between adjacent frames under the relative camera transformation, as defined in Sec.~\ref{sec:video_generation}.

The reconstruction loss $\mathcal{L}_{\mathrm{rec}}$ constrains the 4D representation through differentiable rendering. For each time step $t$, the reconstructed scene is rendered from camera pose $\changed{\mathbf{C}_t}$ and compared with the corresponding generated or target frame:
\begin{equation}\label{eq:rec_loss}
\mathcal{L}_{\mathrm{rec}} =
\sum_{t=1}^{T}
\left\|
\mathcal{R}(\mathcal{S},\changed{\mathbf{C}_t},t)-V_t^{*}
\right\|_1,
\end{equation}
\changed{where $\mathcal{S}$ denotes the reconstructed scene representation, consisting of time-indexed pointmaps in a shared world frame with per-pixel confidence. The pointmaps are predicted by the pose-conditioned CUT3R-based reconstruction module, and $\mathcal{R}(\cdot)$ renders them using differentiable point splatting.}
% where $\mathcal{S}$ denotes the learned 4D scene representation and $\mathcal{R}(\cdot)$ is the differentiable renderer. 
To reduce temporal flickering and encourage smooth scene evolution, we further introduce a temporal regularization term,
\begin{equation}\label{eq:temp_loss}
\begin{split}
\mathcal{L}_{\mathrm{temp}} =
\sum_{t=1}^{T-1} \Big\| &
\Phi(\mathcal{R}(\mathcal{S},\changed{\mathbf{C}_{t+1}},t+1)) \\
& -
\operatorname{Warp}\bigl(
\Phi(\mathcal{R}(\mathcal{S},\changed{\mathbf{C}_t},t)),
F_{t\rightarrow t+1}
\bigr)
\Big\|_1,
\end{split}
\end{equation}
where $\Phi(\cdot)$ denotes an image or feature-space representation, and $F_{t\rightarrow t+1}$ is the optical flow between adjacent frames. This term aligns the reconstructed scene across time and discourages inconsistent geometry or appearance changes that are not supported by the estimated motion.

In practice, we adopt a progressive training schedule. The VLM is first fine-tuned to produce semantically valid camera-motion primitives. The video diffusion module is then optimized with pose-conditioned denoising and geometric consistency losses. Finally, the 4D reconstruction module is trained with rendered-frame supervision and temporal regularization. After these stages converge, we jointly fine-tune the pose encoder, diffusion conditioning layers, and reconstruction adapters while keeping the large pretrained backbones partially frozen. This design avoids unstable gradients through discrete trajectory selection and diffusion sampling, yet still aligns the full pipeline through shared pose conditioning and reconstruction-level supervision. As a result, the generated video is optimized not only for frame-level realism, but also for downstream 4D reconstructability, yielding temporally persistent and geometrically coherent dynamic scenes.

\begin{figure}[!t]
    \centering
    \includegraphics[width=0.93\linewidth]{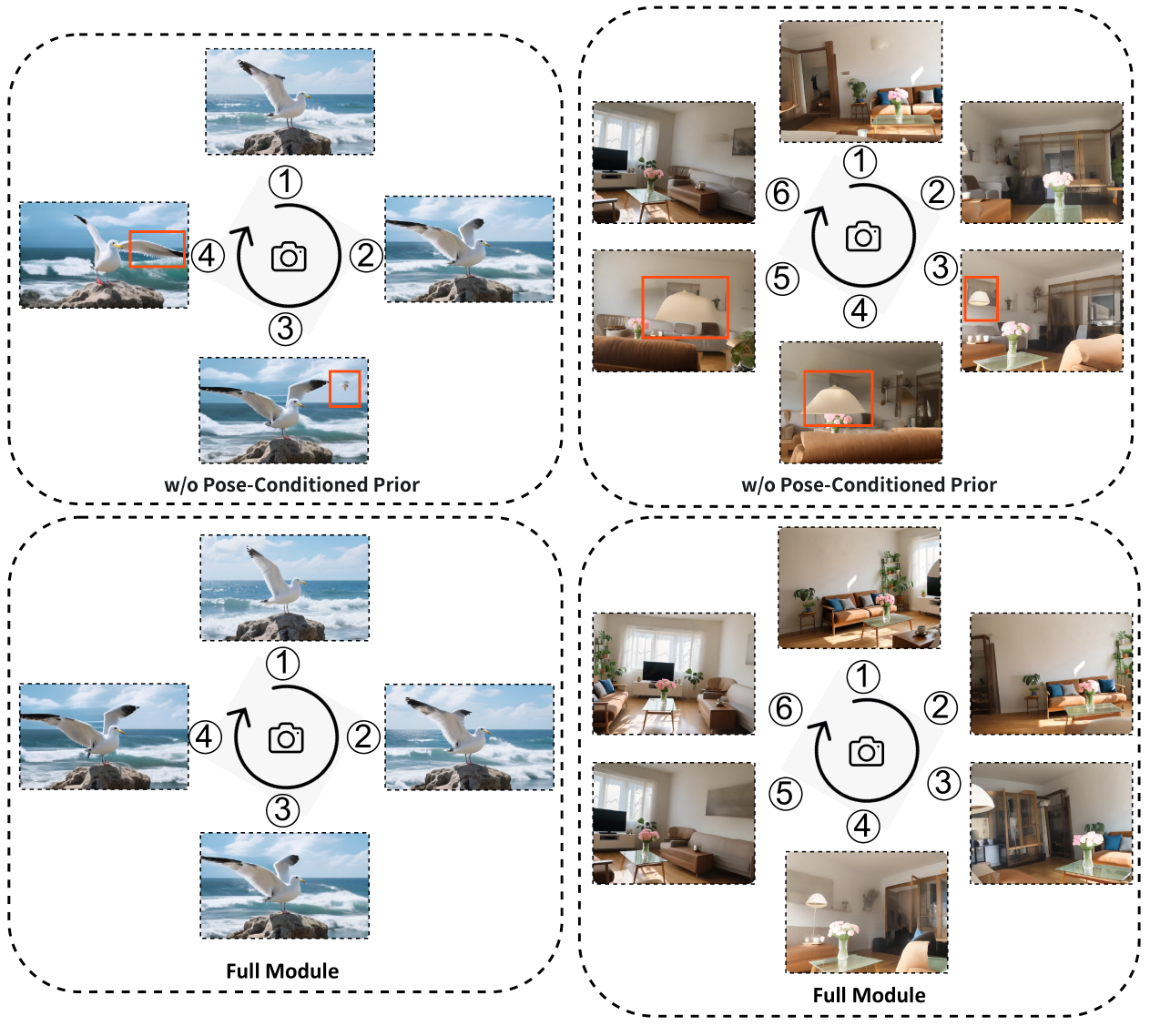}
    \caption{\textbf{Qualitative Ablation Study on the Pose-Conditioned VD Prior.} We compare our full module (with camera guidance) against a baseline that uses a vanilla I2V model without explicit camera control (top row, labeled as 'w/o Pose-Conditioned Prior').}
    \label{fig:ablation}
\end{figure}

\section{Experiments}
\label{section:exp}
% We evaluate our method across a range of 3D tasks, including single and video depth estimation (Sec. 4.1), camera pose estimation (Sec. 4.2), and 3D reconstruction (Sec. 4.3).

% In this section, we introduce the implementation details such as the model architecture, datasets, evaluation metrics, and comparison baselines (see xx Section), as well as the evaluation results included on xx tasks (see xx Section).

In this section, we introduce the details of the implementation of our model architecture and evaluation metrics, as well as the evaluation results included in pose consistency (Sec.~\ref{subsec:vce}), temporal consistency (Sec.~\ref{subsec:tce}), ablation study (Sec.~\ref{subsec:as}) and reconstruction quality (Sec.~\ref{subsec:rq}).
% Together, these results validate the effectiveness of our unified design in maintaining spatiotemporal coherence, particularly under complex occlusions and non-rigid motions.

% \vspace{-\baselineskip}

\subsection{Implementation Details}
\label{subsec:ID}

% \textbf{Baseline}. We compare Our methods with recent camera pose estimation methods on a.

% \noindent
% \textbf{Dataset}. Our methodology integrates large-scale public datasets, specifically RealEstate10K~\cite{realestate} for spatial scene understanding and COCO~\cite{coco} for object-level annotation, alongside AIGC-generated synthetic images to address potential gaps in real-world data coverage. 

% Public datasets provide foundational visual diversity and annotation reliability, while the AIGC component enables controlled augmentation for challenging scenarios (e.g., rare viewpoints or lighting conditions). Rigorous filtering ensures the compatibility of the dataset, and all data are partitioned to maintain a clear separation between the training, validation, and testing phases. These datasets are carefully curated to balance realism, diversity, and scalability, ensuring comprehensive evaluation and reliable performance in different scenarios.

\subsubsection{Architecture.}
% We employ Qwen3-VL technical report~\cite{yang2025qwen3vl} as the base model and finetune it using LoRA~\cite{hu2022lora}. The training dataset consists of 300 image samples manually curated from open-world datasets, each annotated with its optimal camera motion category. The finetuning process requires approximately 2 hours on a single NVIDIA A100 GPU.
We employ Qwen3-VL~\cite{yang2025qwen3vl} as the base model and finetune it on 300 image samples manually curated from open-world datasets, each annotated with its optimal camera motion category. The finetuning process requires approximately 2 hours on a single NVIDIA A100 GPU.
Our video diffusion module is built upon the CamI2V~\cite{cami2v} architecture, which takes an input reference image and a sequence of camera pose parameters as conditions and outputs a dynamic video sequence that adheres to the specified camera pose trajectory.
% Building upon this foundation, we propose a fine-tuning strategy based on a fixed trajectory prior. 
To finetune this module, we employ a dataset comprising 2,000 pose-annotated sequences from RealEstate10K~\cite{realestate} and 1,000 synthetic videos. Synthetic sequences are generated by applying standard camera trajectories with high viewing-angle coverage to diverse images containing static and dynamic scenes.
% The advantage lies in the fact that by guiding the viewpoint through substantial spatial transformations around the scene, the generation model is compelled to infer and render visual information not directly covered in the initial frame, thereby achieving more comprehensive semantic reconstruction of the scene along the temporal dimension.
% Our 4D generator is based on the CUT3R architecture~\cite{cut3r} and achieves precise control over dynamic generation by introducing a camera pose conditioning module. Specifically, we encode the input pose trajectory through an MLP and inject it into the core workflow of CUT3R.
\changed{Our 4D generator builds on CUT3R~\cite{cut3r} with the pose-conditioning pathway described in Sec.~\ref{sec:4dgen}.}
\changed{Detailed curation criteria are provided in Appendix~\ref{app:data}.}

%This condition vector interacts with the visual features extracted by the model from the image sequence and the persistent state vector maintained internally, dynamically adjusting the 3D scene reconstruction process at each time step, thereby generating a 4D dynamic sequence that matches the specified camera trajectory and has a high degree of spatiotemporal consistency.

% Our video generation pipeline decomposes the synthesis process into static and dynamic components: the Static Module constructs consistent multi-view scenes through 3D-aware diffusion, while the Dynamic Module injects controlled camera motion via trajectory-guided epipolar attention, enabling decoupled control over scene content and viewpoint dynamics. We then provide camera motion prompts to fix the optimal camera pose and split the videos generated by the two pipelines into video frames as input for the 4D generator, achieving 4D reconstruction in diverse scenarios.

% cut3r architecture intro

% \noindent precise control over dynamic generation by introducing a camera pose conditioning module. Specifically, we encode the input pose trajectory through a MLP and inject it into the core workflow of CUT3R.

\begin{figure*}[t]
    \centering
    \includegraphics[width=0.91\linewidth]{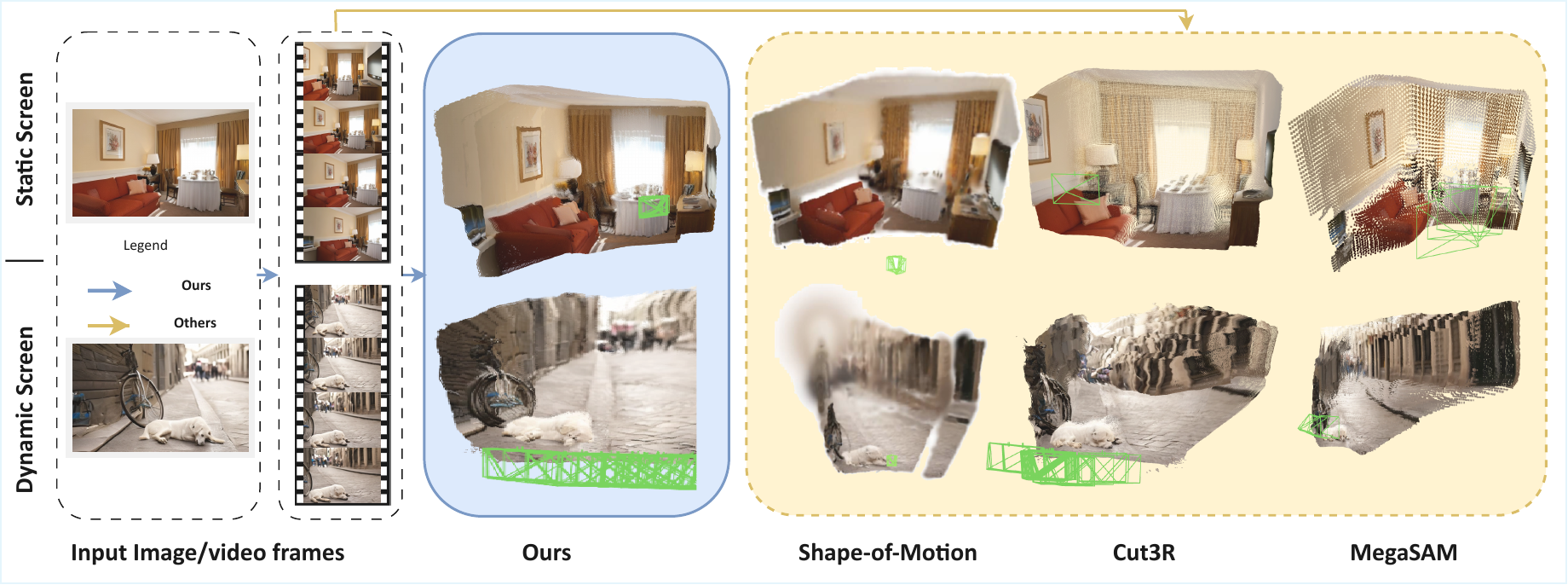}
    \caption{\textbf{Qualitative Results.} We compare our method with concurrent works Shape-of-Motion~\cite{wang2024shape}, Cut3R~\cite{cut3r} and MegaSAM~\cite{li2024_megasam}. Our method achieves the best qualitative results.}
    \label{fig:qualitive}
\end{figure*}

\subsubsection{Metrics.}

In terms of assessing pose consistency, we adopt the Umeyama alignment methodology~\cite{umeyama2002least}, which involves computing a Sim(3) transformation to optimally align the estimated and ground truth trajectories globally, as in \cite{chen2024leap, zhang2024monst3r}. Three standard error metrics are used: the Absolute Trajectory Error (ATE), the Relative Translation Error (RTE), and the Relative Rotation Error (RRE). Regarding temporal consistency evaluation, we build three well-established image quality assessment metrics between consecutive frames throughout the entire sequence: Mean Peak Signal-to-Noise Ratio (mPSNR), Mean Structural Similarity Index (mSSIM) and Mean Learned Perceptual Image Patch Similarity (mLPIPS).

To comprehensively evaluate the impact of the VLM module on video generation quality, we additionally employed several CLIP-based metrics alongside selected dimensions from VBench~\cite{huang2023vbench}. Specifically, we use CLIP-I for image fidelity, CLIP-T for text alignment. To provide a multi-faceted assessment, we employ the Imaging Quality (IQ), Temporal Flickering (TF), and Aesthetic Quality (AQ) metrics from the VBench suite.

% \noindent Quality (AQ) metrics from the VBench suite.

\subsection{Pose Consistency}
\label{subsec:vce}
Following CUT3R~\cite{cut3r}, we compare our method with other state-of-the-art methods on Sintel~\cite{butler2012naturalistic}, TUM-dynamics~\cite{sturm2012benchmark}, ScanNet~\cite{dai2017scannet} and Bonn~\cite{palazzolo2019refusion} datasets to evaluate pose consistency. Specifically, we primarily utilized the complex motion synthetic sequences from the Sintel dataset and the dynamic scene sequences from the TUM-dynamics dataset to evaluate the robustness of various methods in realistic dynamic environments. For the ScanNet dataset, while employing its multiple indoor scene sequences, we supplemented them with dynamic indoor scene sequences from the Bonn Dataset. 

\changed{Dream4D reconstructs from frames synthesized along the VLM-planned trajectory, whereas baselines process
real captured sequences. All methods are evaluated against the same
ground truth with identical Sim(3) alignment and metric definitions.
Since synthesized frames are strictly noisier inputs, Tab.~\ref{tab:table8} serves as
a pose-consistency diagnostic rather than a strictly input-matched
comparison.}

% \noindent the ScanNet dataset, while employing its multiple indoor scene sequences, we supplemented them with dynamic indoor scene sequences from the Bonn Dataset. 

% To evaluate pose consistency, we adopt the Umeyama alignment methodology~\cite{umeyama2002least}, which involves computing a Sim(3) transformation to optimally align the estimated and ground truth trajectories globally, as in \cite{chen2024leap, zhang2024monst3r}. Three standard error metrics are used: the Absolute Trajectory Error (ATE), the Relative Translation Error (RTE), and the Relative Rotation Error (RRE).
As shown in Tab.~\ref{tab:table8}, on the Sintel dataset, although there is a gap in our ATE compared to the state-of-the-art optimization method MonST3R by a \changed{margin of 0.099}, it is already superior to other optimization methods such as DUSt3R (exceeds a significant absolute difference of 0.207). Furthermore, our RRE (0.619) achieves the best level among \changed{online} methods. The performance of our method on the TUM-dynamic dataset is within the ATE of 0.006 and the RTE of 0.002 of the state-of-the-art MASt3R method. The experimental results demonstrate that our proposed method achieves highly competitive performance on the vast majority of metrics, significantly surpassing online methods and matching the accuracy of optimization approaches that require complex iterations.
% As shown in \cref{tab:table8}, our method demonstrates significant improvements and achieves competitive or even superior results across all error metrics on the benchmark datasets. In particular, the video sequences in the Sintel and TUM-dynamics datasets contain complex dynamic objects, which pose certain challenges for viewpoint estimation in traditional methods. As a result, performance across various metrics in these datasets is somewhat affected.
In addition to quantitative analysis, Fig.~\ref{fig:dynamic-demo} demonstrates the point cloud mapping process from a non-fixed perspective, highlighting dynamic environmental interactions.

% \begin{table}[t]
%     \centering
%     \tablestyle{2pt}{1.1}
%     \caption{\textbf{Evaluation on Temporal Consistency.} Our Method achieves superior performance over other benchmarks.}
%     \label{tab:comparison}
%     \begin{tabular}{lcccccc}
%     \toprule
%     \textbf{Method} & \textbf{mPSNR(dB)}$\uparrow$ & \textbf{mSSIM}$\uparrow$ &     \textbf{mLPIPS}$\downarrow$  \\
%     \midrule
%     MegaSAM~\cite{li2024_megasam} & 17.625 & 0.601 & 0.207  \\
%     Shape-of-Motion~\cite{wang2024shape} & 16.68 & 0.630 & 0.332  \\
%     % Cami2V & 14.08 & 0.449 & 0.334  \\
%     Cut3R~\cite{cut3r} & 14.69 & 0.543 & 0.341  \\
%     DynIBaR~\cite{li2023dynibar} & 12.67 & 0.495 & 0.579  \\
%     \rowcolor{gray!20}
%     \textbf{Ours} & \textbf{20.56} & \textbf{0.702} & \textbf{0.170}  \\
%     \bottomrule
%     \end{tabular}
% \end{table}

%\newpage
\subsection{Temporal Consistency}
\label{subsec:tce}
% To quantitatively assess the temporal consistency of 4D sequences, we innovatively employ a frame-to-frame comparison approach, calculating three well-established image quality assessment metrics (mPSNR, mSSIM, and mLPIPS) between consecutive frames throughout the sequence.

% To quantitatively assess the temporal consistency of 4D sequences, we employ a frame-to-frame comparison approach on a benchmark on temporal consistency. 
We employ a frame-to-frame comparison approach on a temporal consistency benchmark to quantitatively assess 4D temporal consistency. 
% , calculating three well-established image quality assessment metrics between consecutive frames across the entire sequence: Mean Peak Signal-to-Noise Ratio (mPSNR), Mean Structural Similarity Index (mSSIM), and Mean Learned Perceptual Image Patch Similarity (mLPIPS). (This methodological framework allows for a comprehensive analysis of 4D coherence across pixel-level, structural, and perceptual dimensions).
To mitigate the risk of temporal consistency metrics artificially rewarding overly static outputs, our evaluation sequences purposefully encompass a diverse range of dynamics, including both subtle movements and large-scale complex motions.
As presented in Tab.~\ref{tab:comparison}, our method represents a 16.6\% improvement over MegaSAM  and a 23.0\% improvement over Shape-of-Motion. This is complemented by a leading mSSIM of 0.702, representing a relative improvement of approximately 11.4\% over Shape-of-Motion and 29.3\% over Cut3R, indicating excellent structural preservation. The performance advantage is equally pronounced in the mLPIPS metric. Our approach's value of 0.170 is only 50\% of the value achieved by Shape-of-Motion.
% presents a comprehensive quantitative comparison of four state-of-the-art 4D reconstruction approaches, evaluated according to the metrics mentioned above. 
The results demonstrate significant variations in the performance of temporal consistency among different methods, with our approach achieving superior scores on all measures, highlighting the effectiveness of our proposed architecture.
% This systematic benchmarking not only highlights the effectiveness of our proposed architecture, but also reveals critical insights into the strengths and limitations of different methodological paradigms in handling spatiotemporal reconstruction tasks. 
% The following \cref{sec:analysis} will examine these comparative results in detail, focusing on the technical innovations that contribute to these performance differences.
To illustrate temporal coherence, Fig.~\ref{fig:static-demo} visually traces the mapping results from a fixed perspective, highlighting the evolution of the scene over time.

\subsection{Ablation Study}
\label{subsec:as}
To validate the effectiveness of the modules in our pipeline, we perform ablation studies on the VLM-based trajectory planning and the spatiotemporal priors from the VD. We further validate where pose conditioning takes effect in Appendix~\ref{ACPC}. Removing only the
reconstruction-side pose cross-attention raises the Sintel ATE from
0.210 to 0.225, and the cascade variant performs worst overall.

% \vspace{-1mm}
\subsubsection{VLM module.}

As shown in Tab.~\ref{tab:ablation}, when the VD module replaces the VLM conditions without trajectory (Empty), there is a significant drop in performance in all metrics: mPSNR decreases by 4.09 dB, mSSIM drops by 0.1972, and the CLIP-T and CLIP-I scores decline by 0.0211 and 0.0231, respectively. In contrast, replacing the VLM with a random trajectory (Random) results in moderate degradation, with mPSNR dropping by 3.7 dB and mSSIM by 0.039, while mLPIPS increases slightly due to reduced structural fidelity.
The results confirm that the VLM contributes to both higher semantic alignment fidelity and better temporal coherence in the generated videos.
These results demonstrate that the VLM not only enhances perceptual and semantic alignment but also contributes substantially to overall image quality, with its absence leading to substantial performance loss.

Notably, the "Random" variant serves as an extreme stress test for VLM planning errors. Despite the measurable performance drop, the system does not collapse; it still reconstructs a coherent (albeit degraded) 4D scene. This provides empirical evidence that, while accurate VLM planning is essential for optimal quality, subsequent stages inherently possess error-compensation capabilities, which prevents pipeline breakdown under incorrect trajectory input.

As presented in Tab.~\ref{tab:ablation4}, the ablation analysis on temporal consistency metrics further underscores the importance of the VLM. Removing the trajectory input (Empty) leads to a sharp decline in IQ by 13.5\%, TF by 3.3\% and AQ by 14.5\%, indicating degraded temporal coherence and motion smoothness. The Random variant shows even more severe deterioration, especially in IQ (down to 67.47), suggesting that random embeddings fail to capture meaningful temporal dynamics. Consistent performance degradation across all variants without VLM highlights its essential role in maintaining high-quality temporal consistency.
% , confirming that our architecture’s strength lies in leveraging language-guided visual understanding for robust spatiotemporal modeling.

\begin{figure*}[t]
    \centering
    \includegraphics[width=0.85\linewidth]{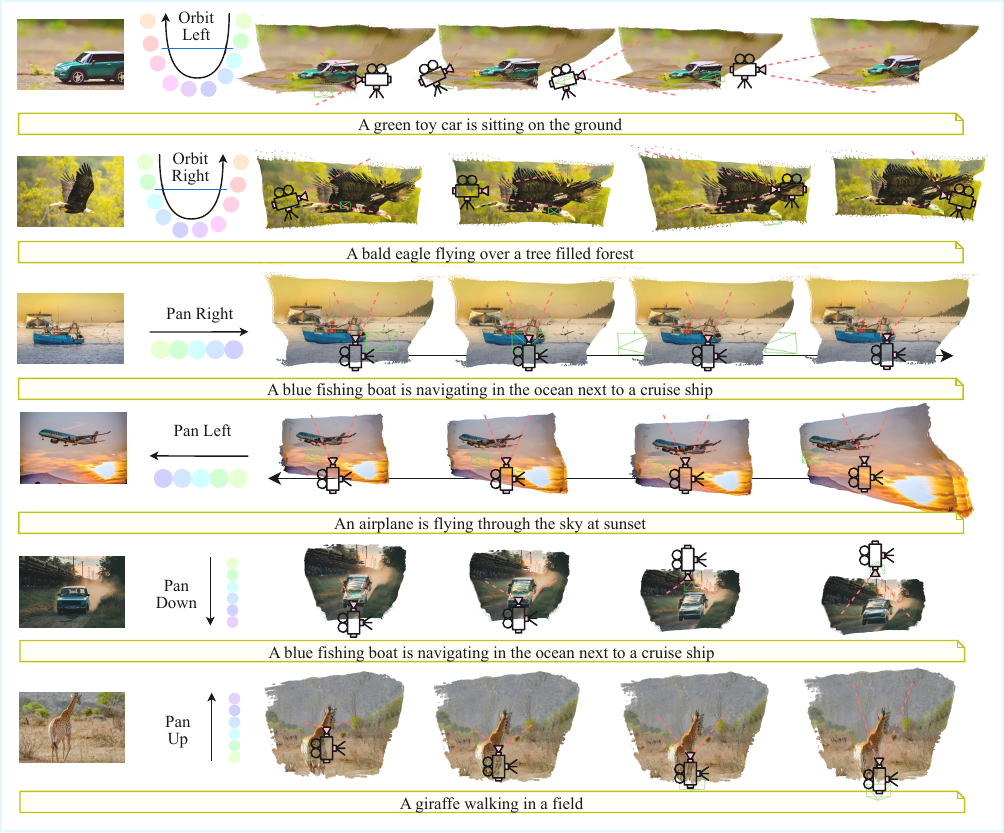}
    \caption{\textbf{Qualitative Results with Diverse Camera Trajectories.} We demonstrate the controllability and robustness of Dream4D across various camera movements, including Orbit (Left/Right) and Pan (Right/Left/Down/Up). For each example, the leftmost column shows the single input image, the middle column visualizes the specified 3D camera trajectory (color-coded from start to end), and the right sequence displays the generated 4D point clouds.}
    \label{fig:supp}
\end{figure*}

% \vspace{-1.5mm}
\subsubsection{Video diffusion model.} %To evaluate the contribution of the spatiotemporal priors from the VD, we divide the finetuning dataset into dynamic and static subsets.
To assess the necessity of world knowledge priors, we retrain the VD on dynamic and static subsets of our finetuning dataset (described in Appendix.~\ref{app:data}).
% We retrain the VD on dynamic and static subsets from the finetuning dataset to assess the necessity of incorporating world knowledge priors.
``Dynamic Prior'' refers to the spatiotemporal knowledge extracted from the video diffusion model fine-tuned on dynamic scene subsets, while ``Static Prior'' refers to the model fine-tuned only on static multi-view data.
For a fair comparison, the ablated versions are trained on an equal number of samples that include dynamic and static scenes.

The quantitative experimental results of the ablation study are presented in Tab.~\ref{tab:ablation2}. The complete approach with dynamic prior achieves the best performance on all metrics, surpassing its ablated variant by significant margins (+1.41 dB in mPSNR, +0.0325 in mSSIM and -0.0621 in mLPIPS). It demonstrates the synergistic effect of integrating dynamic generation with 4D reconstruction. Replacing the VLM conditions with random trajectory or no trajectory inputs for VD leads to significant degradation, which demonstrates the VLM module's ability to resolve temporal inconsistencies that pure diffusion-based generation cannot eliminate alone. 

Beyond quantitative comparisons, Fig.~\ref{fig:ablation} visually demonstrates the critical role of the VD module in achieving spatiotemporal consistency. In a dynamic outdoor scene (upper left), this method generates a discontinuous seagull motion, while in a structurally complex indoor environment (upper right), it struggles to maintain geometric stability during viewpoint changes. Our VD module (bottom row) significantly improves both: enabling smooth bird flight paths (bottom left) and coherent geometry (bottom right). In particular, it simultaneously ensures (1) temporal smoothness (seagull sequence \ding{172} $\rightarrow$ \ding{173} $\rightarrow$ \ding{174}) and (2) spatial integrity (consistent furniture across views \ding{172} $\rightarrow$ \ding{175}). By bridging discrete 3D reconstruction with continuous 4D modeling, the VD module creates a synergistic effect that resolves temporal inconsistencies unattainable by pure diffusion methods.

%5 and 6

\vspace{-2mm}
\subsection{Reconstruction Quality}
\label{subsec:rq}
% \cref{fig:dynamic-demo} and ~\cref{fig:static-demo} demonstrate the point cloud mapping effects from non-fixed and fixed perspectives respectively, showcasing dynamic environmental interactions and temporal evolution of scenes.

%\cref{fig:qualitive} presents a visual comparison of four state-of-the-art 4D reconstruction methods in static and dynamic scenarios. The upper ``Static Screen" section demonstrates performance on indoor scene reconstruction (sofa, furniture), while the lower ``Dynamic Screen" section evaluates temporal consistency in street-view sequences. Each panel contrasts the input frames with the corresponding reconstruction output, where the green lens represents different camera poses. This side-by-side visualization reveals our method's superior geometric accuracy and temporal coherence, particularly in preserving structural details and handling dynamic elements.
Fig.~\ref{fig:qualitive} presents a visual comparison of four state-of-the-art 4D reconstruction methods in static and dynamic scenarios. The upper ``Static Screen'' section demonstrates performance in indoor scene reconstruction (sofa, furniture), while the lower ``Dynamic Screen'' section evaluates temporal consistency in street-view sequences. Each panel contrasts the input frames with the corresponding reconstruction output, where the green lens represents different camera poses. 

% The visual comparison validates our method's superior performance across multiple dimensions. It demonstrates enhanced geometric accuracy in static scenes and remarkable temporal consistency in dynamic sequences, handling motion blur and maintaining dynamic element coherence. This provides strong evidence for the quantitative metrics, highlighting our method's capability to generate spatiotemporally coherent 4D scene representations. Crucially, while competing approaches often sacrifice geometric accuracy for temporal smoothness, Dream4D enables synergistic improvements across both dimensions, ensuring spatiotemporal coherence without compromising structural integrity.

To further validate the effectiveness of our explicit camera control mechanism, we present additional visual results in Fig.~\ref{fig:supp}. The figure illustrates 4D scenes generated from single images under a diverse set of predefined camera trajectories, ranging from orbital rotations to linear panning along different axes. As shown in the visualizations, Dream4D successfully synthesizes high-fidelity novel views that strictly follow the intended camera paths. Notably, the generated scenes exhibit excellent pose consistency, with static backgrounds (e.g., the forest floor or the ocean horizon) remaining geometrically stable, and temporal consistency, where dynamic subjects (e.g., the flying eagle or the moving vehicle) evolve smoothly over time.  These results affirm the model's capability to handle complex camera-subject interactions beyond simple static scene reconstruction. \changed{Beyond automated metrics, a blind pairwise user study with 32 participants
over 24 scenes (Appendix~\ref{app:userstudy}) further confirms the
perceptual advantage of Dream4D.}

\section{Discussion}

While Dream4D achieves robust scene-level consistency, generative quality degrades under extreme trajectories involving extensive translations or large rotations. This stems from compounded constraints in VLM spatial reasoning and diffusion models' long-horizon motion modeling. We emphasize this reflects an engineering ceiling rather than a structural flaw. Introducing metric-scale 3D grounding offers a direct remedy: enabling physically grounded trajectory planning and yielding reconstructions with interpretable spatial metrics.

\vspace{-2mm}
We propose three extension directions. First, semi-supervised trajectory prediction to generalize beyond VLM's single-pass inference for novel motion patterns. Second, integrating physical constraints---such as rigid-body dynamics or material elasticity---into spatiotemporal generation to enhance kinematic plausibility. Third, multi-sensor fusion incorporating stereo vision or depth sensors to strengthen geometric constraints under challenging conditions.

\vspace{-2mm}
Dream4D's modular architecture inherently facilitates these advances. Standardized intermediate representations permit independent component upgrades: advanced vision-language models for richer interaction, or reinforcement learning for adaptive trajectory optimization. This plug-and-play design positions Dream4D as an extensible platform for sustained 4D generation research.

\section{Conclusion}

We propose Dream4D, a novel framework unifying video generation and 4D reconstruction via explicit camera control. Its three-stage pipeline employs a VLM for context-aligned trajectory planning, a pose-conditioned diffusion model for geometrically consistent video synthesis, and a 4D reconstruction module for coherent spatiotemporal recovery. These stages respectively manage semantic understanding, visual generation, and geometric constraints, seamlessly bridging high-level semantics and low-level geometry via standardized representations. Comprehensive evaluations demonstrate superior pose and temporal consistency, yielding geometrically stable and visually compelling dynamic scenes. Ultimately, this unified approach establishes a novel paradigm for comprehending and synthesizing high-quality dynamic scenes.

%-------------------------------------------------------------------------
% bibtex
\bibliographystyle{eg-alpha-doi} 
\bibliography{egbibsample}       

% biblatex with biber
% \printbibliography                

%-------------------------------------------------------------------------
%Color tables are no longer required for purely electronic publications.

\ifincludeappendix
\clearpage
\appendix

\section{Details on Pose-Conditioned 4D Reconstruction}
\label{app:A}

The final stage involves reconstructing a coherent 4D representation from the generated video. We use a spatiotemporal feature field $\mathbf{Q}(\mathbf{x}, t)$ based on a Transformer architecture, as shown in Equ.~\ref{equ:6}. The 3D coordinates $\mathbf{x}$ are first mapped to a higher-dimensional space using Rotational Position Encoding $\gamma(\mathbf{x})$ to capture geometric relationships.

% \begin{equation}
%     \gamma(p) = (..., sin(2^L \pi p), cos(2^L \pi p), ...).
% \end{equation}

% \begin{equation}
% \begin{aligned}
% f(x_m)_{2i-1:2i} = \begin{pmatrix} \cos(m\theta_i) & -\sin(m\theta_i) \\ \sin(m\theta_i) & \cos(m\theta_i) \end{pmatrix} \kern-0.5em \begin{pmatrix} x_{m, 2i-1} \\ x_{m, 2i} \end{pmatrix}.
% \end{aligned}
% \end{equation}

Let $R(\phi)$ denote a 2D rotation matrix by an angle $\phi$, then the transformation can be more compactly expressed as:
\begin{equation}
f(x_m)_{2i-1:2i} = R(m\theta_i) \begin{pmatrix} x_{m, 2i-1} \\ x_{m, 2i} \end{pmatrix}.
\end{equation}
Here, $x_m$ is the input feature vector at position $m$,
% The rotation angle $m\theta_i$ depends on both the position $m$ and the feature dimension $i$, where 
the base rotation angle $\theta_i = 10000^{-2i/d}$, and $d$ is the feature dimensionality.

% \begin{equation}
%     X_t = \{ \mathbf{C}_t^{-1} \cdot \Pi^{-1}(u, v, D_t(u, v)) \mid (u, v) \in \Omega \}.
% \end{equation}

We leverage cross-attention~\cite{vaswani2017attention} for integrating pose information, which correlates the camera pose $\mathbf{C}_t$ with the geometric and appearance features.
This mechanism can be formulated as

\begin{equation}
    Attention(Q_f, K_p, V_p) = Softmax(\frac{Q_f K_p^T}{\sqrt{d_k}}) V_p.
    \label{equ:10}
\end{equation}
Here, $Q_f$ represents query features derived from spatiotemporal features $(\gamma(x), h(t))$, while $K_p$ and $V_p$ are key and value features derived from the pose embedding of $\mathbf{C}_t$.

\section{Implementation Details}
\label{app:implementation}

\subsection{Data Curation}
\label{app:data}

\changed{The 300 VLM fine-tuning examples are sampled from OpenImages~\cite{openimages} and
COCO~\cite{coco} to cover diverse indoor and outdoor scenes. Each example is
annotated by the authors with a scene-appropriate motion primitive,
including motion type, direction, and coarse magnitude, using a
predefined cinematography rubric. The resulting set contains
60 stationary, 120 translation, and 120 rotation examples.
For video-diffusion fine-tuning, we retain 2,000 RealEstate10K~\cite{realestate} clips
with complete camera poses and at least 30 frames, all verified by manual inspection. Using the camera-motion-compensated residual
optical flow, these clips are divided into 1,200 static-dominant and
800 motion-rich sequences. For the controlled comparison in Tab.~\ref{tab:ablation2},
800 clips are sampled from each subset to keep the training-set size
identical.
The 1,000 synthetic videos are generated from OpenImages~\cite{openimages} samples by
depth-based view synthesis: monocular depth is estimated to lift each
image into 3D, novel views along the sampled trajectory are rendered,
and disoccluded regions are inpainted. Source images are disjoint from the
evaluation set.}

\subsection{Training Configurations}
Our training pipeline consists of three stages: VLM-based trajectory planning, pose-conditioned video generation, and 4D reconstruction.
Tab.~\ref{tab:training_config} summarizes the hyperparameters used in training and generation pipelines.
We use these configurations across experiments, unless otherwise noted.

For VLM-based trajectory planning, we employ Qwen3-VL~\cite{yang2025qwen3vl} as the base model.
We utilize LoRA~\cite{hu2022lora} fine-tuning to achieve efficient parameter tuning.
% To ensure that the generated 4D content aligns with the user's semantic intention, we employ a Vision-Language Model (VLM) to infer appropriate camera movements. As described in \cref{sec:trajectory_selection}, the VLM acts as a high-level planner that translates the input image and the text description into a sequence of executable camera poses. 
We detail the specific prompt structure designed to guide the VLM below:
% ,  with a rank of $r=8$ and $\alpha = 16$. The model is optimized using the AdamW optimizer ($\beta_1=0.9, \beta_2=0.95$) with a weight decay of 0.01. We set the learning rate to $1\times 10^{-4}$ and train for 5 epochs with a batch size of 8.

\begin{table}[t]
\centering\footnotesize
\caption{\textbf{Summary of training hyperparameters}}
\label{tab:training_config}
\begin{tabular*}{\linewidth}{@{\extracolsep{\fill}}ll}
\toprule
\textbf{Parameter} & \textbf{Value} \\
\midrule
% --- Classification Backbone ---
\multicolumn{2}{l}{\textit{VLM Fine-Tuning}} \\
LoRA rank ($r$) & 8 \\
LoRA alpha ($\alpha$) & 16 \\
Optimizer & AdamW ($\beta_1=0.9, \beta_2=0.95$) \\
Learning rate & $1 \times 10^{-4}$ \\
Weight Decay & 0.01 \\
Batch Size & 8 \\
Epochs & 5 \\
% Warmup Steps & 20 \\
% Hardware
\midrule
% --- VD Trainging ---
\multicolumn{2}{l}{\textit{Configuration for Video Diffusion Model}} \\
% Resolution & $256 \times 256$ \\
% Frame Count ($T$) & 16 frames \\
Optimizer & AdamW ($\beta_1=0.9, \beta_2=0.95$) \\
Learning rate & $1 \times 10^{-4}$ \\
Weight decay & 0.01 \\
Batch size & 48 \\
% Mixed Precision & 16-mixed (FP16) \\
Training steps & 30,000 \\
Noise schedule & Linear ($\beta$: $8.5\text{e-}4 \to 1.2\text{e-}2$) \\
% ($\beta_{start} = 8.5 \times 10^{-4}, \beta_{end} = 1.2 \times 10^{-2}$)
% Hardware & 6 $\times$ RTX 4090 (24GB) \\

% Camera Trace Scale Factor & 2-4 \\
% Trace Extract Ratio & 0.3-0.4 \\
% Text CFG Scale & 6 \\
% Sampling Steps & 40 \\
% Random Seed & 12333 \\
\midrule
% --- 4D Generator Training ---
\multicolumn{2}{l}{\textit{Optimization Settings for 4D Reconstruction}} \\
% Initial resolution & 256$\times$256 \\
% Final resolution & 512$\times$512 \\
Optimizer & AdamW ($\beta_1=0.9, \beta_2=0.95$) \\
Learning rate & $1 \times 10^{-6}$ \\
Batch size & 6 \\
% Initial learning rate & $1 \times 10^{-6}$ \\
% Final learning rate & $1 \times 10^{-7}$ \\
% Warmup Epochs & 0.5 \\
Weight decay & 0.05 \\
Epochs & 10 \\
Depth estimator    & \changed{DepthAnything-V2 (frozen)} \\
Flow estimator     & \changed{UniMatch (frozen)} \\
\midrule
\multicolumn{2}{@{}l}{\emph{Joint Objective (Eq.~\ref{eq:total_loss_compact})}} \\
$\lambda_{\text{plan}}$ / $\lambda_{\text{DM}}$ / $\lambda_{\text{rec}}$ & \changed{1.0 / 1.0 / 1.0} \\
$\lambda_{\text{geo}}$ / $\lambda_{\text{temp}}$ & \changed{0.1 / 0.1} \\
% Warm-up duration (epochs) & 0.5 \\
% Checkpoint saving frequency & every 0.1 epoch \\
% Validation frequency & every 1 epoch \\
\bottomrule
\end{tabular*}
\end{table}

\begin{list}{}{\leftmargin=0.8em \rightmargin=0.8em}
  \item\relax
\textit{``You are an experienced on-set cinematographer directing a crucial scene. Your primary task is to determine the next camera movement instruction based on the current frame and its description, relying on your professional intuition. You must carefully analyze the user's description of the current shot and apply expert cinematographic thinking to select the most logical subsequent movement from three types of camera motions.
\begin{itemize}
\item Static: Camera keeps the current angle with no movement.
\item Translation: Camera performs a pan/tilt move. Always specify direction (up, down, left, right) and motion amplitude (subtle/moderate/wide sweep). Example: "moderate pan to the left".
\item Rotation: Camera rotates around the optical center. Always specify direction (up, down, left, right) and rotation amplitude (subtle/moderate/wide sweep). Example: "subtle upward rotation.
\end{itemize}
Please note that you only output the predicted shooting trajectory of the cameramen for the next step. Except for static shooting, specific directions need to be provided for translation and rotation''
}    
\end{list}
% \begin{tcolorbox}[title=\textbf{System Prompt}, fontupper=\small\ttfamily]
% \#\#\# Task \\
% You are an experienced on-site photographer and are directing the shooting of a key scene. Your current core task is to determine the next camera movement command based on the current scene and description, relying on your professional intuition. 
% \end{tcolorbox}

% The video diffusion model is trained to generate clips with a resolution of $256 \times 256$ and a sequence length of $T=16$ frames. We utilize 6 NVIDIA RTX 4090 GPU (24GB VRAM), setting a global batch size of 48 (8 per GPU) and employing 16-bit mixed precision (FP16). The model is trained for 30,000 steps with a learning rate of $1 \times 10^{-4}$ and a linear noise schedule ranging from $\beta_{start} = 8.5 \times 10^{-4}$ to $\beta_{end} = 1.2 \times 10^{-2}$. 

During the video diffusion inference process, we use a Classifier-Free Guidance (CFG) scale of 6.0 and perform 40 sampling steps. Specifically for our trajectory control, the camera trace scale factor is set between 2 and 4, with a trace extraction ratio of 0.3 to 0.4. The trace-scale factor is applied after primitive decoding to control the total trajectory extent.

% For the final 4D generation stage, we adopt a coarse-to-fine optimization strategy. The rendering resolution starts at $256 \times 256$ and progressively increases to $512 \times 512$. To ensure fine-grained convergence, the learning rate is annealed from an initial value of $1 \times 10^{-5}$ to $1 \times 10^{-6}$.

\section{Discussion on Camera Control and Dual-Conditioning Strategy}
\label{app:camera_control}

Our motion categories function as parameterized primitives rather than rigid constraints, enabling compositional and fine-grained camera control. In the video generation stage, Dream4D employs a dual-conditioning strategy to achieve both precision and flexibility. Specifically, the VD model receives explicit $SE(3)$ parameters mapped from the VLM to ensure geometric precision adherence to the camera trajectory. Simultaneously, the VD model leverages the textual prompt for implicit semantic guidance of scene dynamics. \changed{Consequently, our framework supports continuous camera trajectories
within the supported primitive family, composed from parameterized
motion primitives with continuous amplitudes, while maintaining high
fidelity to the user's instructions.}

\begin{table}[t]
    \centering
    \caption{\textbf{Ablation Studies Across Multiple Dimensions of VLM on VBench.}
    % Dream4D’s robust performance on VBench highlights the VLM's critical role in enhancing video generation. By excelling in Subject Consistency (SC) for identity preservation, Background Consistency (BC) for environmental stability, and Motion Smoothness (MS) for natural movement, the model demonstrates superior spatiotemporal quality. These evaluations jointly validate the effectiveness of our VLM integration. 
    SC: Subject Consistency. BC: Background Consistency. MS: Motion Smoothness. }
    \label{tab:ablation5}
    \tablestyle{15pt}{1.2}
    \begin{tabular}{lccc}
    \toprule
    \textbf{Variant} & \textbf{SC$\uparrow$} & \textbf{BC$\uparrow$} & \textbf{MS$\uparrow$} \\
    \midrule
    w/ VLM  & \textbf{0.9762} & \textbf{0.9783} & \textbf{0.9932} \\
    \rowcolor{gray!10}
    Random  & 0.9658 & 0.9640 & 0.9845 \\
    \rowcolor{gray!20}
    $\Delta$ & -0.0104 & -0.0143 & -0.0087 \\
    \rowcolor{gray!10}
    Empty  & 0.9578 & 0.9577 & 0.9757 \\
    \rowcolor{gray!20}
    $\Delta$ &  -0.0184 &  -0.0206 &  -0.0175 \\
    \bottomrule
    \end{tabular}
\end{table}

\section{Explanation and Expansion of Evaluation Metrics}
To rigorously evaluate our method's performance, we employ a comprehensive set of metrics targeting different aspects of generation quality: pose accuracy, temporal stability, and overall video quality. The following sections explain why each metric was chosen and supplement with additional VBench~\cite{huang2023vbench} metrics. \changed{We further conducted a blind pairwise study with 32 participants on 24 held-out scenes.}

\subsection{Why We Use These Pose Consistency Metrics}

Dream4D's core contribution is generating spatiotemporally consistent 4D scenes via explicit camera control.  Consequently, the quality of our final output is fundamentally dependent on the geometric accuracy of the camera trajectory predicted by the VLM. Since a temporally coherent 4D scene must inherently satisfy strict 3D geometric constraints, traditional frame-by-frame visual metrics are insufficient. Therefore, we adopt the 3D reconstruction metric (ATE, RTE, RRE) to directly measure camera path correctness, a direct proxy for scene's geometric integrity.

\subsection{Why We Use These Temporal Consistency Metrics}

Standard temporal consistency metrics typically rely on optical flow estimation to align frames. However, these methods often suffer from error propagation, where artifacts in generated videos mislead the flow estimator, rendering the metric unreliable. Furthermore, they often fail to decouple high-frequency flickering from natural motion dynamics.

To address these limitations, we adopt a robust, multi-level frame-to-frame comparison framework. Instead of relying on potentially noisy flow priors, we directly evaluate the coherence of adjacent frames. This approach is grounded in the principle that valid temporal changes (camera motion) differ fundamentally from artifacts (flickering) in feature space.  By assessing similarity across a spectrum of features, from low-level pixel fidelity to high-level perceptual semantics, our metric provides a more discriminative and holistic assessment of temporal stability.

\subsection{Why We Use CLIP and VBench Metrics}

% Finally, we evaluate our method with the VBench suite to ensure that its results are not only technically sound, but also of high quality and aesthetic appeal.

% We use CLIP scores to confirm that the generated video is semantically correct. CLIP-I verifies that the video content remains faithful to the initial source image, while CLIP-T ensures that it aligns with the user's text prompt. 

% We also report Imaging Quality (IQ), Temporal Flickering (TF), and Aesthetic Quality (AQ) from VBench. These metrics provide a standardized benchmark for overall video quality and stability.

To verify the effectiveness of our video generation module, we conduct a comprehensive evaluation using the CLIP-based metrics and VBench suite. 
To assess semantic correctness, we employ CLIP-based metrics: CLIP-I measures faithfulness to the initial image, while CLIP-T evaluates alignment with the text prompt.  Beyond semantic accuracy, we also report standardized scores for Imaging Quality (IQ), Temporal Flickering (TF), and Aesthetic Quality (AQ) from VBench.

\section{Ablation on Cross-Stage Pose Conditioning}
\label{ACPC}

Tab.~\ref{tab:ablation_pose} analyzes where camera-pose conditioning takes effect.
All three variants consume the same generated sequences, so the
differences isolate the reconstruction side. Removing only the
reconstruction-side pose cross-attention raises ATE from 0.210 to
0.225 and lowers mPSNR from 20.56 dB to 19.63 dB. This confirms
that pose guidance contributes directly to reconstruction quality,
beyond its role in video generation. The cascade variant, which
replaces our reconstruction adaptation with the original CUT3R,
degrades further to an ATE of 0.233 and 18.92 dB. The gap between these two variants shows that the reconstruction adaptation
helps even without pose conditioning, since it adapts CUT3R to
the statistics of generated videos. Notably, the full model attains an
ATE of 0.210 on generated inputs, matching CUT3R on real captured frames (0.213 in Tab. 1). This indicates that pose conditioning
compensates for the additional noise in synthesized observations.

% \begin{figure*}[h]
%     \centering
%     \includegraphics[width=\linewidth]{./Supplementary.pdf}
%     \caption{\textbf{Additional Qualitative Results with Diverse Camera Trajectories.} We demonstrate the controllability and robustness of Dream4D across various camera movements, including Orbit (Left/Right) and Pan (Right/Left/Down/Up). For each example, the leftmost column shows the single input image, the middle column visualizes the specified 3D camera trajectory (color-coded from start to end), and the right sequence displays the generated 4D point clouds. The results highlight our method's ability to maintain spatiotemporal consistency and geometric stability while adhering to both the input image and the text instructions.}
%     \label{fig:supp}
% \end{figure*}

% \subsection{Computational Cost}

% \begin{table}[htbp]
% \centering
% \caption{Pipeline Stage Resource and Time Summary}
% \label{tab:pipeline_summary}
% \begin{tabular}{l r r}
% \toprule
% \textbf{Pipeline Stage} & \textbf{Params / VRAM} & \textbf{Inference Time (per scene)} \\
% \midrule
% Stage 1: Trajectory Planning & $\sim$7B / 14GB & $\sim$ 2 seconds \\
% Stage 2: Video Generation & $\sim$2B / 18GB & $\sim$ 1.5 minutes \\
% Stage 3: 4D Reconstruction & - / 22GB & $\sim$ 15 minutes \\
% \midrule
% Total Pipeline & Max VRAM: $\sim$22GB & $\sim$ 17 minutes \\
% \bottomrule
% \end{tabular}
% \end{table}

\begin{table}[t]
  \centering
  \caption{\textbf{Results of the blind pairwise user study on the final 4D
    renderings.} Each cell reports the percentage of trials in which
    participants preferred our method (O), rated both as tied (T), or
    preferred the baseline (B). Geo.\ Cons.\ denotes geometric
    consistency, Temp.\ Smooth.\ denotes temporal smoothness, and
    Overall 4D denotes overall 4D generation quality.}
  \label{tab:user_study}
  \resizebox{\columnwidth}{!}{%
  \rowcolors{2}{gray!8}{white}
  \begin{tabular}{@{}lccc@{}}
    \toprule
    Comparison & \shortstack{Geo.\\Cons.} & \shortstack{Temp.\\Smooth.} & \shortstack{Overall\\4D} \\
    \midrule
    Ours vs.\ CUT3R           & \textbf{74.2}\,/\,8.6\,/\,17.2  & \textbf{78.1}\,/\,7.0\,/\,14.8 & \textbf{75.8}\,/\,7.8\,/\,16.4 \\
    Ours vs.\ MegaSAM         & \textbf{66.4}\,/\,10.9\,/\,22.7 & \textbf{64.8}\,/\,10.2\,/\,25.0 & \textbf{65.6}\,/\,10.2\,/\,24.2 \\
    Ours vs.\ Shape-of-Motion & \textbf{64.8}\,/\,10.5\,/\,24.6 & \textbf{69.1}\,/\,9.4\,/\,21.5 & \textbf{68.0}\,/\,9.8\,/\,22.3 \\
    \bottomrule
  \end{tabular}%
  }
\end{table}

\begin{table}[t]
\centering
\caption{\textbf{Ablation of cross-stage camera-pose conditioning.} Pose errors are evaluated on Sintel, while mPSNR is evaluated on the temporal-consistency benchmark. Cascade feeds the CamI2V-generated sequence directly into the original CUT3R. w/o Rec. Pose retains our reconstruction adaptation but removes only the reconstruction-side pose cross-attention. All three variants use the same generated observations.}
\label{tab:ablation_pose}
\footnotesize
\setlength{\tabcolsep}{2pt}
\resizebox{\columnwidth}{!}{%
\begin{tabular}{@{}lccc cccc@{}}
\toprule
Variant & \shortstack{Pose\\in VD} & \shortstack{Rec.\\Adapt.} & \shortstack{Pose\\Cond.} & ATE$\downarrow$ & RTE$\downarrow$ & RRE$\downarrow$ & mPSNR$\uparrow$ \\
\midrule
CamI2V$\rightarrow$CUT3R & \checkmark & $\times$ & $\times$ & 0.233 & 0.064 & 0.674 & 18.92 \\
w/o Rec. Pose & \checkmark & \checkmark & $\times$ & 0.225 & 0.061 & 0.658 & 19.63 \\
\textbf{Full Dream4D} & \checkmark & \checkmark & \checkmark & \textbf{0.210} & \textbf{0.054} & \textbf{0.619} & \textbf{20.56} \\
\bottomrule
\end{tabular}%
}
\end{table}

\subsection{Additional Vbench metrics results}

To provide a more comprehensive assessment beyond the results in the main text, we conduct additional experiments specifically focusing on Video Quality. We evaluate Subject Consistency (SC), Background Consistency (BC) and Motion Smoothness (MS) with the VBench benchmark. It is worth noting that the unreported VBench metrics primarily target stylistic consistency or semantic consistency, which are less critical for verifying our core contribution of spatiotemporal coherence and geometric stability.

Tab.~\ref{tab:ablation5} demonstrates the superior spatiotemporal quality achieved by the integration of VLM. The 'w/ VLM' model surpasses the 'Empty' baseline by significant margins, delivering relative improvements of approximately 1.92\% in Subject Consistency and 2.1\% in Background Consistency, alongside a 1.9\% gain in Motion Smoothness. The comparison with the 'Random' variant further highlights that the VLM offers meaningful semantic guidance (e.g., -0.0143 in BC). This confirms that the VLM is pivotal in ensuring temporally consistent and visually coherent video generation.

\subsection{Additional User study results}
\label{app:userstudy}

\changed{To assess perceptual quality beyond automated metrics, we conducted a
blind pairwise user study comparing Dream4D against CUT3R, MegaSAM, and
Shape-of-Motion. 32 participants each viewed 24 generated 4D scenes.
For every pair, they chose the better result in terms of geometric
consistency, temporal smoothness, and overall 4D quality.
The presentation order and the left--right placement were randomized.
Ties were allowed.
As shown in Tab.~\ref{tab:user_study}, Dream4D is preferred in
64.8\% to 78.1\% of all pairwise comparisons, while ties remain below 11\%,
indicating that the gains are decisive rather than marginal. The advantage is most pronounced against our backbone CUT3R, where
Dream4D is chosen 4.3$\times$ to 5.3$\times$ more often
(74.2\% to 78.1\% vs.\ 14.8\% to 17.2\%). }

\fi

\end{document}